\documentclass[lettersize,journal]{IEEEtran}
\usepackage{amsmath,amsfonts}
\usepackage{algorithmic}
\usepackage{algorithm}
\usepackage{array}
\usepackage[caption=false,font=normalsize,labelfont=sf,textfont=sf]{subfig}
\usepackage{textcomp}
\usepackage{stfloats}
\usepackage{url}
\usepackage{verbatim}
\usepackage{graphicx}
\usepackage{cite}
\usepackage{booktabs}  
\usepackage[table]{xcolor} 
\usepackage{hyperref} 
\usepackage{multicol}

\usepackage{url}            
\usepackage{booktabs}       
\usepackage{multirow}    
\usepackage{amsfonts}       
\usepackage{nicefrac}       
\usepackage{microtype}      
\usepackage{enumerate}
\usepackage{hhline}
\usepackage{makecell}
\usepackage{pifont}

\usepackage{graphicx} 
\usepackage{caption}
\usepackage{subcaption}
\usepackage{amsmath}
\usepackage{amsthm}
\usepackage{amssymb}
\usepackage{tikz}
\usepackage{xcolor}
\usetikzlibrary{arrows}

\allowdisplaybreaks

\usepackage{mathrsfs}

\usepackage{algorithm}
\usepackage{algorithmic}
\usepackage{hyperref}
\usepackage{bm}

\allowdisplaybreaks

\usepackage[capitalize,noabbrev]{cleveref}
\crefname{thm}{Theorem}{Theorems}
\crefname{lem}{Lemma}{Lemmas}
\crefname{cor}{Corollary}{Corollaries}
\crefname{prop}{Proposition}{Propositions}
\crefname{asmp}{Assumption}{Assumptions}
\crefname{defn}{Definition}{Definitions}
\crefname{oracle}{Oracle}{Oracles}
\crefname{fact}{Fact}{Facts}
\crefname{conj}{Conjecture}{Conjectures}
\crefname{rem}{Remark}{Remarks}
\crefname{claim}{Claim}{Claims}
\crefname{ec}{Empirical Observation}{Empirical Observations}

\definecolor{red}{rgb}{1, 0, 0}

\definecolor{green}{rgb}{0, 1, 0}
\definecolor{darkgreen}{rgb}{0.0, 0.2, 0.13}
\definecolor{darkseagreen}{rgb}{0.56, 0.74, 0.56}
\definecolor{officegreen}{rgb}{0.0, 0.5, 0.0}

\definecolor{blue}{rgb}{0, 0, 1}

\definecolor{orange}{rgb}{1, 0.4, 0.0}

\input{packages/math_commands}

\newcommand{\etal}[0]{~et al.}
\newcommand{\figtext}[1]{{\footnotesize #1}}

\begin{document}

\title{Model Synthesis for Zero-Shot Model Attribution}

\author{
  Tianyun Yang, Juan Cao, Danding Wang, Chang Xu~\IEEEmembership{Member,~IEEE}
  
\thanks{Tianyun Yang, Juan Cao, and Danding Wang are with the Institute of Computing Technology, Chinese Academy of Sciences, Beijing, China. \protect E-mail: \{yangtianyun19z,caojuan,wangdanding\}@ict.ac.cn

Chang Xu is with School of Computer Science, Faculty of Engineering, University of Sydney, Australia. E-mail: c.xu@sydney.edu.au 

Code: \href{https://github.com/TianyunYoung/Model-Fingerprint}{https://github.com/TianyunYoung/Model-Fingerprint}
}}



\maketitle

\begin{abstract}

Nowadays, generative models are shaping various fields such as art, design, and human-computer interaction, yet they are accompanied by copyright infringement and content management challenges. In response, existing research seeks to identify the unique fingerprints on the images they generate, which can be leveraged to attribute the generated images to their source models. However, existing methods are restricted to identifying models within a static set included in classifier training, incapable of adapting dynamically to newly emerging unseen models. To bridge this gap, this paper aims to develop a generalized model fingerprint extractor capable of zero-shot attribution that effectively attributes unseen models without exposure during training. Central to our method is a model synthesis technique, which generates numerous synthetic models that mimic the fingerprint patterns of real-world generative models. The design of the synthesis technique is motivated by observations on how the basic generative model’s architecture building blocks and parameters influence fingerprint patterns, and it is validated through designed metrics to examine synthetic models' fidelity. Our experiments demonstrate that the fingerprint extractor, trained solely on synthetic models, achieves impressive zero-shot generalization on a wide range of real-world generative models, improving model identification and verification accuracy on unseen models by over 40\% and 15\%, respectively, compared to existing approaches.
\end{abstract}


\section{Introduction}
\IEEEPARstart{I}{n} recent years, advanced generative (vision) models have revolutionized various fields such as art creation, design, and human-computer interaction~\cite{nichol2021glide, ramesh2022hierarchical, rombach2022high, mj}. Despite their positive impact, these models have also given rise to new concerns, such as copyright infringement issues and content supervision. To address these concerns, model attribution, the process of identifying the source model of generated content, has gained increasing attention \cite{marra2019gans,yu2019attributing,xuan2019scalable,yang2022aaai,bui2022repmix, yang2023progressive}. It helps deter unauthorized copying and distribution, enabling content creators and rights holders to prove ownership and take legal action against infringements. Furthermore, model attribution allows regulators to identify and act against entities using generative models for harmful, illegal, or unethical purposes. 

\begin{figure*}[th]
    \centering
    \includegraphics[width=0.75\linewidth]{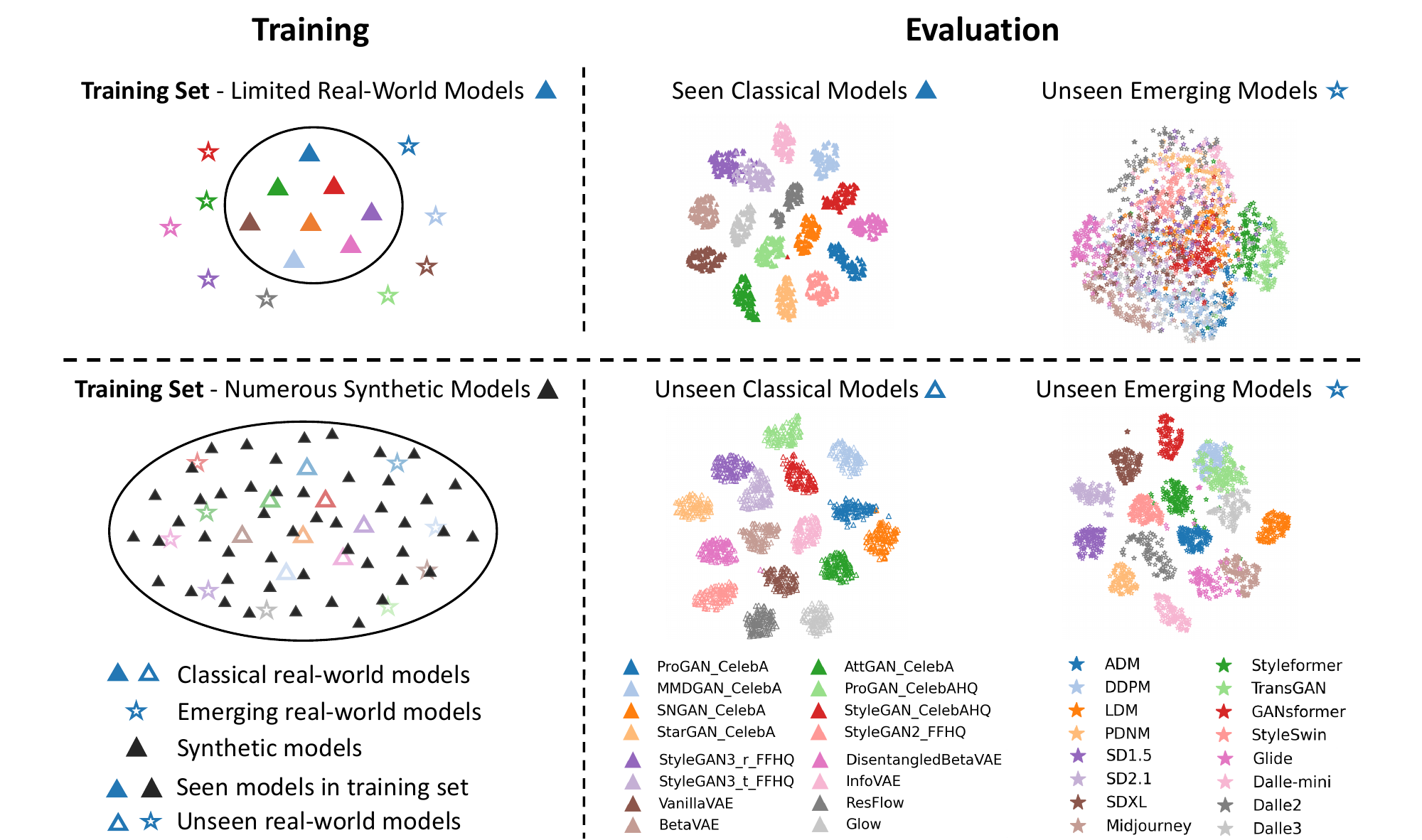}
    \caption{Illustration of the existing methods (above) and ours (below). Existing methods depend on training with a limited set of real-world models, struggling to distinguish emerging unseen models like Stable Diffusion (SD) and DALL-E series. In contrast, our method trains on numerous synthetic models that replicate a broader range model fingerprints, enabling it to distinguish and attribute unseen models effectively.}
    \label{fig:intro}
\end{figure*}



Model attribution methods identify unique fingerprints in images generated by source models. These methods can leverage these fingerprint features to trace a generated image back to its originating model. A commonly adopted model attribution paradigm frames the task as a multiclass classification problem~\cite{yu2019attributing, yang2022aaai, bui2022repmix, yang2023progressive}. In this setup, images generated by a limited and static set of models are used to train a classifier, where each image is labeled with a unique model ID. During testing, a test image comes from a seen model in the training set, and the classifier would identify the model ID based on the model fingerprint it has learned~\cite{yu2019attributing, yang2022aaai, bui2022repmix}. This framework looks promising, but has a major limitation: classical methods fail to attribute models not present in the training data, assigning incorrect classification labels to such new models. This drawback is unacceptable in today's rapidly evolving landscape, where new models emerge quickly. Although some works~\cite{yang2023progressive,abady2024siamese} consider an open-set setup by assigning an ``unknown" label for new models, these approaches still struggle to adapt to the newly emerged unseen models dynamically.


To tackle the flexibility challenge, we aim to develop a more generalized model attribution method capable of zero-shot model attribution, efficiently attributing unseen models \textit{without requiring any training on samples from these models}. Instead of relying on a classification head to classify the generated image to a static set of seen models that have been included in training, we view model attribution as comparing the "fingerprint" distance between the test sample and each model's fingerprint within a model gallery. This framework allows us to easily deal with newly emerged models, we could simply collect samples from new models, use the fingerprint extractor to create their fingerprints, and dynamically updates to the model gallery, without requiring any training on these models.

How to train an effective fingerprint extractor capable of generalizing on unseen models? Traditional wisdom suggests training it on a limited set of real-world models. However, this approach suffers from poor generalization on unseen models, which is exemplified in Figure~\ref{fig:intro} (above). To overcome this, we propose pre-training the extractor with a significantly larger set of synthetic models (e.g., 100 times more in our experiments), which provides similar fingerprint patterns as real-world models but much more diverse. 


Our model synthesis strategy is based on analysis in the \emph{frequency domain}, which is inspired by previous works~\cite{frank2020leveraging, corvi2023intriguing} demonstrating that fingerprints are unobservable in the spatial domain but prominent in the frequency domain. To understand how unique fingerprint patterns of real-world generative models are formed, we examine the mechanism from the perspective of basic building blocks of generative models. Our findings reveal that:
\begin{itemize}
    \item The type of basic network components—such as upsampling, activation functions, normalization, and convolution layer parameters—significantly influences the spectral patterns of generated images.
    \item Due to the upsampling layer naturally attenuating high-frequency components from earlier layers, the last few generative blocks within a large model are more influential in determining the output's spectral patterns.
\end{itemize}
Based on these insights, we hypothesize that synthesizing shallow generative blocks with various combinations of basic components can effectively represent the fingerprint space of unseen models. Therefore, we develop a simple yet effective model synthesise strategy. 

Inspired by the above analysis, we construct the synthetic models based on a shallow auto-encoder architecture, comprising fewer generative blocks compared to typical generative models. We increase synthesis diversity by varying the basis components, including the types of upsampling layers, activation functions, normalization layers, the number and sequence of these layers, and training seeds. Consequently, we get 5760 synthetic models across 288 different architectures by minimizing the reconstruction loss. This model synthesis strategy has advantages in terms of \emph{fidelity} and \emph{efficiency}. By comparing the spectrum distribution between synthetic and real-world models, our results indicate that our synthesis options allow the spectrum distribution of synthetic models to more closely align with those of real-world generative models. Further, synthesizing a model takes only 83 seconds on average with a single 3090 GPU. This is remarkable faster than the naive solution of model synthesis (i.e., training state-of-the-art generative models from scratch). For example, it takes 14 days 22 hours to train a StyleGAN~\cite{karras2019style} model that generate with a 256$\times$256 resolution\footnote{https://github.com/NVlabs/stylegan}, which is significantly slower than ours.

We leverage these synthetic models to train the fingerprint extractor, to extract fingerprints from these models, and distinguish between each other. We combine classification and metric loss to enhance the discrimination of learned fingerprint embedding. Experimental results demonstrate that: 1) Our fingerprint extractor, despite being solely trained on synthetic models, exhibits strong zero-shot attribution capabilities on a broad scope generative models, including the classical GAN, VAE, Flow models, and emerging Diffusion models like Stable Diffusion and DalleE-3. Partial visualization results are shown in Figure~\ref{fig:intro} (below). 2) We consider two model attribution scenarios, including model identification and model verification. Our method significantly outperforms existing approaches on unseen models, achieving accuracy improvements of over 40\% and 15\% respectively. 3) Our method is capable of tracing the LoRA-variants back to their base models only relying on the generated images, providing an efficient defense tool against model intellectual property infringements.


Our major contributions can be summarized as follows:

1) We propose to solve the problem of zero-shot model attribution, which expands the model attribution target to unseen models that are not included in training.

2) We propose solving the zero-shot model attribution problem by training on numerous synthetic models that mimic the fingerprint patterns of real-world generative models. Our synthesis strategy is inspired by observations on how generative model architecture building blocks and parameters influence fingerprint patterns. This strategy has advantages in terms of fidelity and efficiency.

3) Experimental results demonstrate that our fingerprint extractor, trained on synthetic models, exhibits strong generalization capabilities on a wide range of real-world generative models. Compared with existing methods, we improve the model identification and verification accuracy on unseen models by over 40\% and 15\% respectively.

\section{Related work}
\label{sec:related_work}

\subsection{Model Attribution}
\label{subsec:generative_model_attribution}

In essence, model attribution methods rely on some relationship between the internal states of generative models and the generated images. This paper mainly focuses on fingerprint-based model attribution, which relies on the unique  fingerprint patterns on the output images of generative models. Marra et al. \cite{marra2019gans} used averaged noise residuals to represent model fingerprints and found that these fingerprints are periodic. Subsequent works \cite{yu2019attributing, yang2022aaai, bui2022repmix} confirmed the existence of these fingerprints and achieved high accuracy with a fixed, finite set of models following a closed-set classification approach. Asnani et al. \cite{asnani2023reverse} extended this research to trace the architectural components of models that generate images, which is beyond the scope of this paper. In real-world scenarios, images often originate from unseen models not present during training. Recent works \cite{yang2023progressive, abady2024siamese} addressed this by using an open-set approach, which attributes seen models and rejects unseen models not included in training. However, as generative technology continues to evolve, the variety of unseen models expands continuously, challenging the capacity of existing methods to adapt in the real-world dynamically. To bridge this gap, our goal is to develop a more generalized model fingerprint extractor capable of zero-shot model attribution, efficiently attributing unseen models without requiring any training on samples from these models. Table~\ref{tab:method_compare} summarizes the differences between our method and related works mentioned above.

\begin{table*}
\centering
\caption{Differences between our method and related works. Our method enables zero-shot model attribution, allowing us to attribute unseen models without needing training samples from them. In contrast, other methods can only attribute seen models that were included in training.}
\label{tab:method_compare}
\begin{tabular}{ccccc}
\toprule[1.2pt]
Method & Task & Training Data Type & Number of Training Models & Goal \\
\midrule
\cite{marra2019gans,yu2019attributing,yang2022aaai,bui2022repmix} & Closed-set attribution & Real-world models & $<$ 50 & Attribute seen models \\
\cite{yang2023progressive,abady2024siamese} & Open-set attribution & Real-world models & $<$ 50 & Attribute seen models, reject unseen models \\
Ours & Zero-shot attribution & Synthetic models & $>$ 5000 & Attribute unseen models \\
\bottomrule[1.2pt]
\end{tabular}
\end{table*}

\subsection{Spectrum Discrepancies of Generative Models}
\label{subsec:frequency_bias_of_generative_models}

Although fingerprints are visually imperceptible in the spatial domain, they are more noticeable in the frequency domain, often appearing as discrepancies in the spectrum. Existing studies \cite{zhang2019detecting, durall2020watch, schwarz2021frequency, dzanic2020fourier, khayatkhoei2022spatial, chandrasegaran2021closer, corvi2023intriguing} have sought to explain the spectrum discrepancies between generated and real images, particularly by studying the influence of upsampling and convolution layers.


Studies in \cite{zhang2019detecting, durall2020watch, schwarz2021frequency} find that upsampling operations lead to high-frequency discrepancies between real and generated images. Zhang et al. \cite{zhang2019detecting} showed that the upsampling layer causes a periodic grid pattern on the spectrum. Durall et al. \cite{durall2020watch} identified a high-frequency discrepancy caused by the upsampling layer, which makes generative architectures struggle to fit the real image distribution. Schwarz et al. \cite{schwarz2021frequency} discovered that different upsampling operations bias the generator toward distinct spectral properties. Other studies \cite{dzanic2020fourier, khayatkhoei2022spatial} have analyzed the influence of convolution layers and attributed the frequency differences between real and generated images to the linear dependencies in the spectrum of convolutional filters, which hinder the learning of high frequencies.

The studies above primarily focused on differences between real and generated images, with fewer analyzing the frequency discrepancies between images generated by different models. In our study, we further investigate the influence of different types of activation and normalization layers and the effects of various convolution parameters. We have newly observed the influence of different generative blocks, providing additional insights into the formation of model fingerprints.

\subsection{Pre-train on Synthetic Data}


Our study is related to using the synthetic data for improving generalization performance. Baradad\etal~\cite{baradad2021learning} find that diverse noise data with capture certain structural properties of real data, despite far from realistic, could achieve good performance when used for self-supervised learning for a image classification task. Baek\etal~\cite{baek2022commonality} develop the data synthesis strategy by adopting the generic property of the natural images in power spectrum distribution, structure, and existence of saliency. The pretrained GANs using the synthetic dataset can effectively transfer in low-shot adaptation. Mishra\etal~\cite{mishra2022task2sim} propose a task-aware synthesis strategy to find the simulation parameters (lighting, pose, materials, etc.) that best fit the downstream task. 

In contrast to these works, our research proposes synthesizing \emph{models} rather than \emph{images} to mimic the patterns of model fingerprints rather than natural images.

\section{Method}

In this work, our primary goal is to design a fingerprint extractor that could generalize to unseen models in the open world. To accomplish this goal, we propose a new approach by utilizing synthetic models, to mimic the fingerprint patterns of real-world generative models. By this approach, we significantly broaden the scope of the training data of fingerprint extractor, and consequently diminish the generalization gap. In the following, we begin with an analysis of the factors influencing the fingerprint patterns of generative models, detailed in Section~\ref{subsec:warm_up}. Based the analysis, we then design our model synthesis strategy in Section~\ref{subsec:model_synthesis}. Finally, in Section~\ref{subsec:fingerprint_extractor}, we use the synthetic models to train the fingerprint extractor to perform the model attribution task.

\subsection{Preliminary Analysis of Model Fingerprint}
\label{subsec:warm_up}
\begin{figure}
    \centering
    \includegraphics[width=\linewidth]{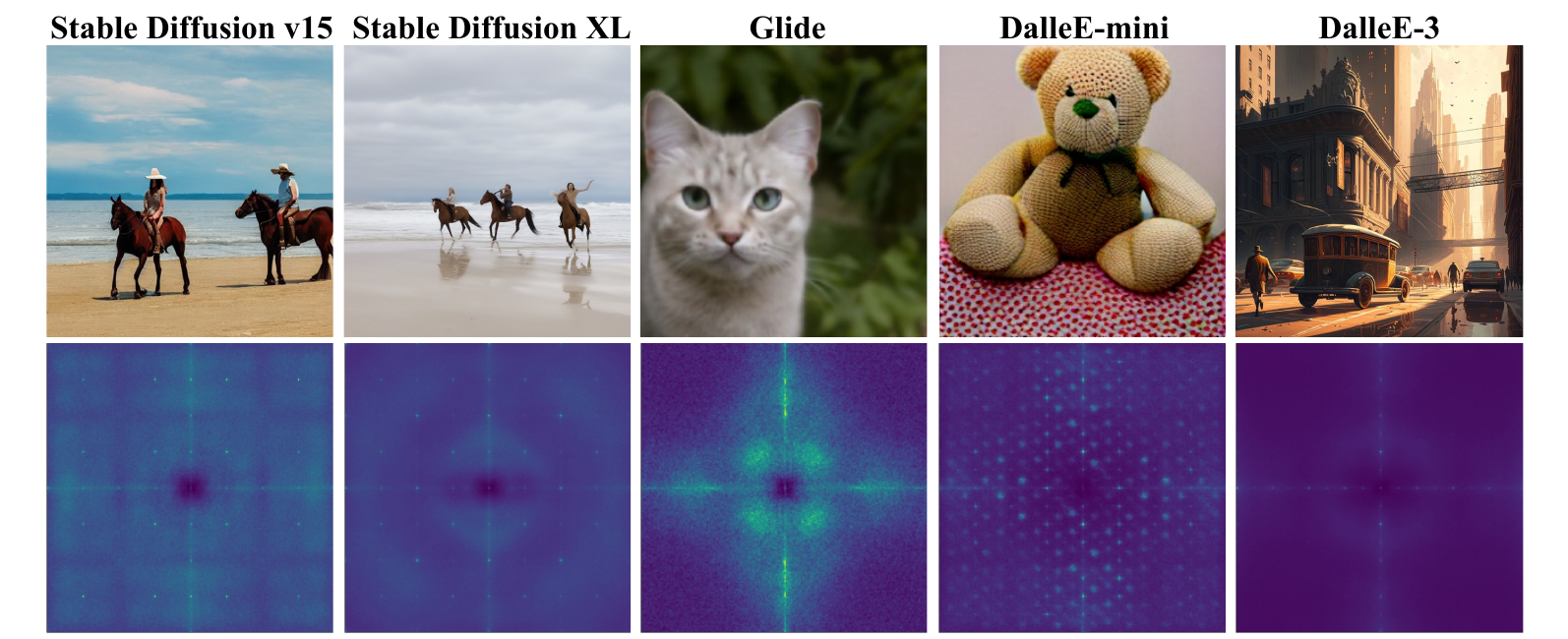}
    \caption{Images (top) produced by generative models and the associated (averaged) Fourier spectrum (bottom). The unique characteristics of the Fourier spectrum can be regarded as the fingerprints of generative models and used for model attribution. The spectrum are obtained from high-pass filtered images processed through a denoiser.}
    \label{fig:spectrums}
\end{figure}

Our study draws inspiration from recent research~\cite{marra2019gans, yu2019attributing, corvi2023intriguing}, which demonstrates that generative models, differing in architecture or parameters, leave unique patterns, termed \emph{model fingerprints}, on their generated images. Despite these fingerprints are typically imperceptible in the spatial domain, they are more evident in the frequency domain. As shown in Figure~\ref{fig:spectrums}, each generative model is characterized by a distinct pattern in the averaged Fourier spectrum. To gain insights for replicating these patterns, we first explore why generative models display unique spectrum patterns. 

The common architecture of an image generative model typically includes several key components designed to generate high-quality images from a latent representation or input data. Some of the most widely used architectures in image generative models include Generative Adversarial Networks (GANs), Variational Autoencoders (VAEs), Diffusion Models, etc. Image generative models, regardless of their specific type, typically consist of a series of generative blocks. Each block is designed to progressively transform the noise/latent vector into an image through several operations. To provide a cohesive mathematical foundation, consider a common generative block architecture consisting of upsampling, convolution, activation, and normalization layers applied to a Gaussian noise input \( \mathbf{z}\). The block's output \( \mathbf{y} \) can be expressed as:

\begin{align}
\mathbf{y} = \mathcal{N}\left( \phi\left( F\left( U\mathbf{z} \right)\right) \right),
\end{align}

where:
\begin{itemize}
    \item \( U \): Upsampling operator.
    \item \( F \): Convolution operator with kernel \( \mathbf{W}_{conv} \) and bias \( \mathbf{b} \).
    \item \( \phi \): Nonlinear activation function.
    \item \( \mathcal{N} \): Normalization function.
\end{itemize}

In the following, we aim to discuss the impact of these basic components and different generative blocks on output's spectrum patterns. To validate the theoretical insights, we also conducted experiments using ProGAN, SNGAN, and StyleGAN2 models with varying network components in Appendix~\ref{sec:appendix_analysis}, analyzing their spectrum influence on output images.

\subsubsection{\textbf{Upsampling Effects}} In the spatial domain, the upsampling operation \( U\textbf{z} \) is equivalant to zero-interleaving the input signal \textbf{z} and then applying a standard convolutional operation, which can be expressed as:
\begin{align}
U \textbf{z} = \text{ZeroInter}(\textbf{z}) \otimes \textbf{K}_{\text{up}},
\end{align}
where \( \text{ZeroInter}(\textbf{z}) \) represents \textit{zero-interleaving} of the input signal \( \textbf{z} \), which inserts zeros between spatial samples. \( \otimes \) denotes the convolution operation in the spatial domain. \( \textbf{K}_{\text{up}} \) represents the upsampling kernel. 

Using the Convolution Theorem, the upsampling operation in the frequency domain becomes:
\begin{align}
\mathcal{F}\{U \textbf{z}\}(\omega) =\text{Repeat}_{2,2}(\mathcal{F}\{\textbf{z}\})(\omega) \cdot \mathcal{F}\{\textbf{K}_{\text{up}}\}(\omega),
\end{align}
where \( \text{Repeat}_{2,2} \) represents the replication of the input spectrum \( \mathcal{F}\{\textbf{z}\} \) along two frequency dimensions by a factor of 2. \( \mathcal{F}\{\textbf{K}_{\text{up}}\} \) represents the Fourier Transform of the upsampling kernel \( \textbf{K}_{\text{up}} \). \( \cdot \) denotes pointwise multiplication in the frequency domain.

\textit{Fingerprint:} The upsampling kernel $\textbf{K}_{\text{up}}$ determines how the replicated spectra are weighted. Its magnitude $|\mathcal{F}\{\textbf{K}_{\text{up}}\}(\omega)|$ directly shapes the output signal. Different types of upsampling are discussed in existing works to leave different spectral properties~\cite{zhang2019detecting, schwarz2021frequency}. Basic kernels like nearest-neighbor or bilinear correspond to low-pass frequency responses, attenuating high-frequency content. In contrast, learned transposed convolution kernels can either preserve, boost, or selectively filter out certain frequency bands. Thus, they tends to preserve more high-frequency components~\cite{zhang2019detecting}. Thus, the upsampling "fingerprint" represents a characteristic spectral pattern formed through frequency shaping by the upsampling kernel.

\subsubsection{\textbf{Convolution Effects}} Applying the Convolution Theorem, convolution \( F \) in the spatial domain translates to multiplication in the frequency domain:
\begin{align}
\mathcal{F}\{ F\mathbf{x} \}(\omega) = \mathcal{F}\{ \mathbf{W}_{conv} \}(\omega) \cdot \mathcal{F}\{ \mathbf{x} \}(\omega) + \mathbf{b} \cdot \delta(\omega),
\end{align}
where $\mathbf{x}$ is the input signal, $\mathbf{W}_{conv}$ and $\mathbf{b}$ are the kernel and bias of the Convolution operator.

\textit{Fingerprint:} The output spectrum is shaped by \( |\mathcal{F}\{ \mathbf{W}_{conv} \}(\omega)|\). Each learned convolution kernel acts as a frequency-selective filter, enhancing certain frequency bands and suppressing others. This creates a spectral “fingerprint” dictated by the learned convolution kernel’s frequency response. 

\subsubsection{\textbf{Nonlinear Activation Effects}} In the spatial domain, applying $\phi(\mathbf{x})$ is elementwise. While in frequency domain, elementwise nonlinearities do not translate to simple pointwise operations. Instead, nonlinearities correspond to convolution-like mixing of frequencies. To simplify the analysis, the nonlinearity can be approximated by a Taylor series expansion. For small values of $\mathbf{x}$, common nonlinear activation functions like ReLU, Sigmoid and Tanh can be approximated using their Taylor series expansions: 
\begin{itemize}
    \item ReLU:  
    The expansion is\footnote{Since ReLU is non-differentiable at $\mathbf{x}=0$, we approximate it using $\text{softplus}_{10}(x)$, and use the Taylor series expansions of $\text{softplus}_{10}(x)$ for ReLU.}:  
    \begin{align}
    ReLU(\mathbf{x}) \approx \frac{\log(2)}{10} + \frac{1}{2}\mathbf{x} - \frac{5}{4}\mathbf{x}^2 + \frac{5}{48}\mathbf{x}^3 + \mathcal{O}(\mathbf{x}^4).
    \end{align}
    \item \text{Tanh}:  
    The expansion is:  
    \begin{align}
    \tanh(\mathbf{x}) \approx \mathbf{x} - \frac{\mathbf{x}^3}{3} + \mathcal{O}(\mathbf{x}^5).
    \end{align}

    \item \text{Sigmoid}:  
    The expansion is:  
    \begin{align}
    \text{sigmoid}(\mathbf{x}) \approx \frac{1}{2} + \frac{\mathbf{x}}{4} - \frac{\mathbf{x}^3}{48} + \mathcal{O}(\mathbf{x}^5).
    \end{align}
\end{itemize}

In the frequency domain, higher-order terms (e.g., $\mathbf{x}^2$) result in repeated convolutions of the input signal's Fourier Transform $H(\omega)$ with itself:
\begin{align}
\mathcal{F}\{\mathbf{x}^n\}(\omega) = \underbrace{H(\omega) * H(\omega) * \cdots * H(\omega)}_{n \text{ times}},
\end{align}
where $H(\omega) = \mathcal{F}\{\mathbf{x}\}(\omega)$. This convolution operation generates harmonics at integer multiples of the input frequency $\omega$. For example, $H(\omega) * H(\omega)$ introduces even-order harmonics, such as $2\omega$, and a DC component ($\omega$=0).

\textit{Fingerprint:} The frequency-domain  “fingerprint” of the nonlinear activation is the introduction of harmonics and frequency mixing patterns, transforming a input into a broader spectral distribution. For instance, ReLU introduces strong even harmonics due to the quadratic term ($\frac{5}{4}\mathbf{x}^2$). Tanh and Sigmoid add weak odd harmonics ($\frac{\mathbf{x}^3}{3}$ and $\frac{\mathbf{x}^3}{48}$).

\subsubsection{\textbf{Normalization Effects}}

Normalization techniques adjust data by normalizing its mean and variance. In the frequency domain, different types of normalization have the same formula as below:
\begin{align}
\mathcal{F}\{\hat{\mathbf{x}}\}(\omega) = \frac{\gamma}{\sigma} \mathcal{F}\{\mathbf{x} - \mu\}(\omega) + \mathcal{F}\{\beta\}(\omega),
\end{align}
where \(\mu\) and \(\sigma\) are the mean and standard deviation. \(\gamma\) and \(\beta\) are learnable scale and shift parameters. \(\mu\) and \(\sigma\) are computed differently depending on the method, such as across the entire batch for Batch Normalization~\cite{ioffe2015batch} or for each individual sample in the case of Instance Normalization~\cite{ulyanov2016instance}. 

\textit{Fingerprint:} Despite having the same frequency formulation, due to the variations in the optimization landscapes they create, different types of normalization can lead to distinctive distributions of parameters~\cite{ulyanov2016instance}, and consequently different spectrum patterns.\footnote{For instance, Figure~\ref{fig:norm_dft} in Appendix~\ref{sec:appendix_analysis} shows that models trained with different seeds but the same normalization type generally exhibit more consistent spectral patterns than those trained with a different normalization method.}

\begin{figure*}
    \centering    
    \includegraphics[width=\linewidth]{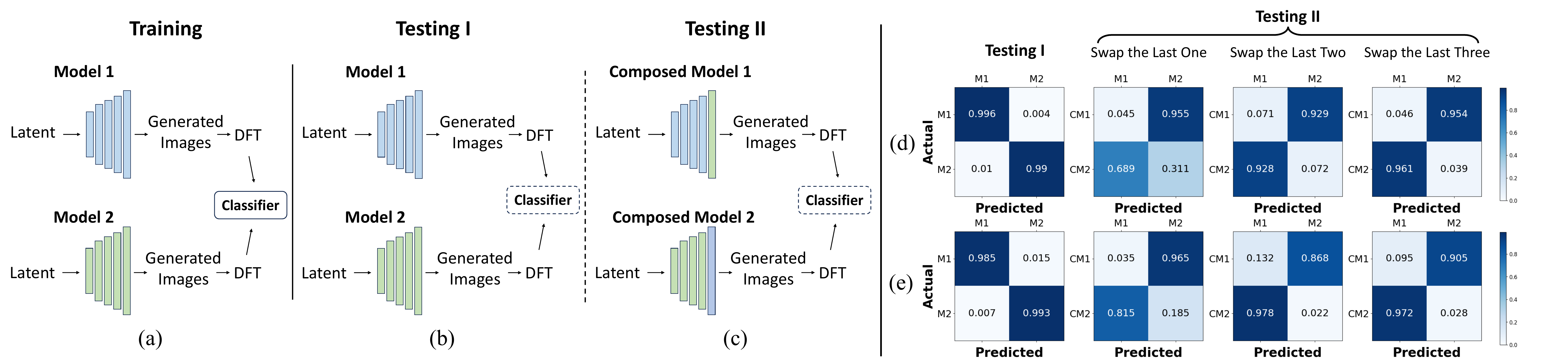}
    \caption{(a-c) Verification of the importance of the last generative block in producing model fingerprints. In training (a), images generated by two models are used to train a 2-way classifier. In the first testing scenario (b), we test on non-overlap testing samples from the two models. In the second testing scenario (c), we test on samples from two composed models, which are constructed by switching the parameters of the last generative blocks of two models in training. (d-e) The confusion matrices in first and second testing scenarios. Model 1 and Model 2 are abbreviated as M1 and M2, respectively. Composed Model 1 and Composed Model 2 are abbreviated as CM1 and CM2.}
    \label{fig:compose}
\end{figure*}
\subsubsection{\textbf{Effects of Different Generative Blocks}} The discussion above focuses on the influence of basic components. Moving forward, it is important to consider how different generative blocks throughout the entire generation process contribute to the final spectral pattern on the generate image. Generative models typically consist of multiple generative blocks, with upsampling layers playing a crucial role as connectors between these blocks, doubling the resolution of the input feature maps. As mentioned above, commonly used upsampling layers, such as bilinear and nearest-neighbor interpolation, inherently possess a low-pass filtering effect. This property leads them to attenuate high-frequency components from prior layers. Given that existing research~\cite{yu2019attributing} suggesting that model fingerprints predominantly reside in the high-frequency components, it becomes apparent that the \textit{later blocks in a model's architecture could more significantly influence the model fingerprints on the output image as the high-frequency patterns from earlier layers could be filtered out.} To empirically test this hypothesis, we designed an experiment depicted in Figure~\ref{fig:compose}, consisting of three key steps:
    \begin{itemize}
    \item \textbf{Training:} Train a classifier on the Fourier spectrum of images generated by two distinct generative models with the same architecture (Model 1 and Model 2). 
    \item \textbf{Testing I:} Measure the classification results using a test set of images generated by Model 1 and Model 2.
    \item \textbf{Testing II:} Starting from the i-th block, sequentially exchange the succeeding blocks from Model 1 to Model 2 and vice versa. After each exchange, generate a new set of images with the composed models and assess the classifier's classification results on this new dataset. Figure~\ref{fig:compose}(c) visualize the operation by exchanging the last block.
    \end{itemize}   
Figure~\ref{fig:compose}(d,e) shows the confusion matrices under Testing I and Testing II of two experiments. We use two ProGAN models for Model 1 and Model 2. In (d), we constructed these models using nearest-neighbor upsampling layers. In (e), the models utilize bilinear upsampling layers. As shown in the Figure, under Testing I, the classifier achieves high accuracy, indicated by the high diagonal values in the confusion matrix. However, after swapping the last few blocks in Testing II, the confusion matrix reveals a notable reversal in model attribution results. Specifically, when swapping the last two blocks, the reversal accuracy reaches 0.9. These shifted results suggest that the most distinguishable spectral patterns for differentiating models are primarily generated by the last two blocks of the models. In contrast, the patterns left by earlier blocks appear to become less distinct as they progress through the generative process.

\subsection{Model Synthesis Strategy}
\begin{figure}[t]
\centering
\includegraphics[width=0.7\linewidth]{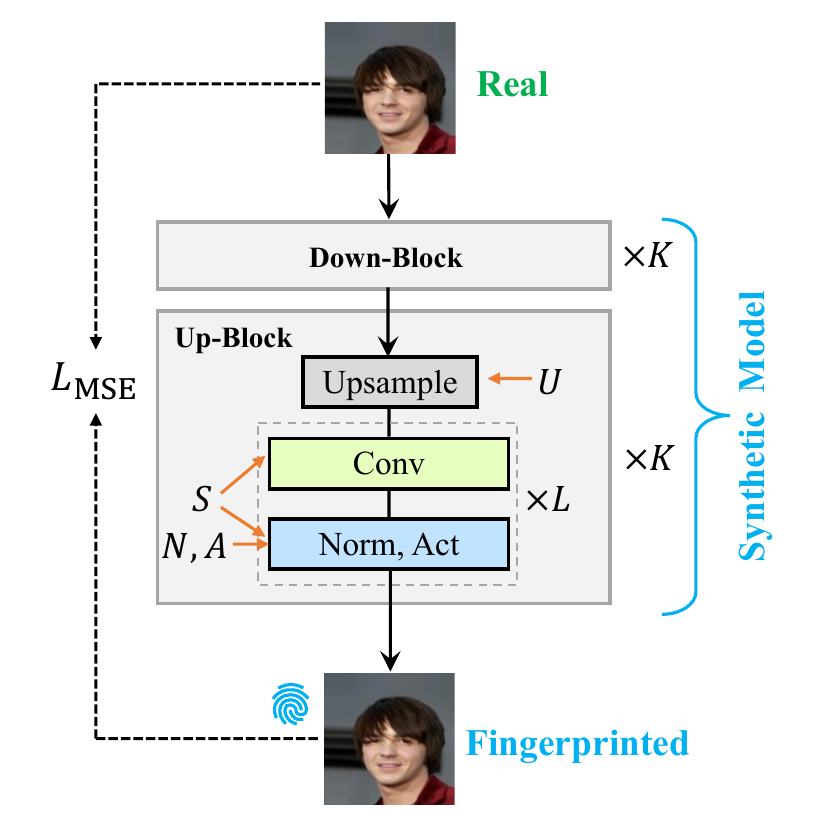}
\caption{Our proposed model synthesis strategy. The synthetic model receives a real image as input, producing a fingerprinted image that contains the distinct fingerprint characteristics associated with the synthetic model.}
\label{fig:generative_blocks}
\end{figure}

\label{subsec:model_synthesis}

In this section, we introduce the model synthesis strategy that we use to create models for training the fingerprint feature extractor. Intuitively, a good synthesis strategy should generate models whose fingerprint patterns closely resemble those of real generative models, allowing these synthetic models to effectively simulate new models in an open environment. According to the discussion in Section~\ref{subsec:warm_up}, there are important factors to consider. First, the last few blocks tend to have a more profound influence on the distinguishable spectrum patterns of the output images. Thus, employing a synthesis architecture with a small number of generative blocks could ensure fidelity in mimicking fingerprints. Second, it is crucial to increase diversity by varying the types of upsampling layers, activation functions, normalization layers, and parameters. We explicitly incorporate these considerations into our method. Moreover, to further enhance diversity, we also consider variations in the number and sequence of layers.  

Overall, our model synthesis strategy is illustrated in Figure~\ref{fig:generative_blocks}, the structure of the synthetic model could be viewed as a shallow auto-encoder, composed of $K$ downsampling blocks and $K$ upsampling blocks. Based on the above discussion, $K$ can selected as a small value, such as one or two. Each downsampling block use a fixed architecture, with a pooling layer to downscale the input resolution by half, and two convolutional layers to increase the feature dimension. The feature output from the downsampling blocks is then sent into the upsampling blocks, which share common components with the generative blocks in standard generative models but offer various architectural options. 

In summary, the architecture of synthetic models is defined by the options $\{K, L, U, A, N, S\}$. 
\begin{itemize}
    \item $K$ refers to the number of downsampling/upsampling blocks within the synthetic model, which can be one or two.
    \item $L$ represents the number of convolution layers per block, which can be one or two.
    \item $U$ is the upsampling operation that can be nearest neighbor upsampling, bilinear upsampling, or a stride 2 transposed convolution layer.
    \item $A$ is the activation function that can be ReLU, Sigmoid, Tanh, or no activation. 
    \item $N$ is the normalization type that can be batch normalization, instance normalization, or no normalization. 
    \item $S$ refers to the order of activation and normalization relative to the convolution layer. 
\end{itemize}
By varying these configurations, we can get $2 \times 2 \times 3 \times 4 \times 3 \times 2 = 288$ different architectures within the construction space. For each architecture, we train $M$ models with different training seeds, creating $M$ distinct models with different parameters per architecture. $M$ is set to 20 in our experiments.

We recognize that speed is a key factor in model synthesis. In our approach, we simplify the objective to focus on reconstruction, where the generative neural network is specifically tasked with minimizing the reconstruction loss of input images. This method is not only straightforward to implement but also allows for rapid training. We leave the exploration of other training approaches to future work. We set a constraint on the minimum reconstruction residual to $\eta = 0.005$ to limit the amount of artifacts in the output images. Ultimately, the total number of synthetic models generated is 5760, allowing for a wide simulation of a wide range of potential generative models with diverse architectures and parameters.

\subsection{Effectiveness of Model Synthesis}
\label{sec:sec43}

We evaluate the effectiveness of our model synthesis strategy in terms of \textit{Fidelity} and \textit{Efficiency}. Fidelity assesses how well the synthetic models replicate the spectrum/fingerprint patterns of real-world generative models. Efficiency examines the speed of training on a large number of synthetic models.

\begin{figure}
    \centering
    \includegraphics[width=\linewidth]{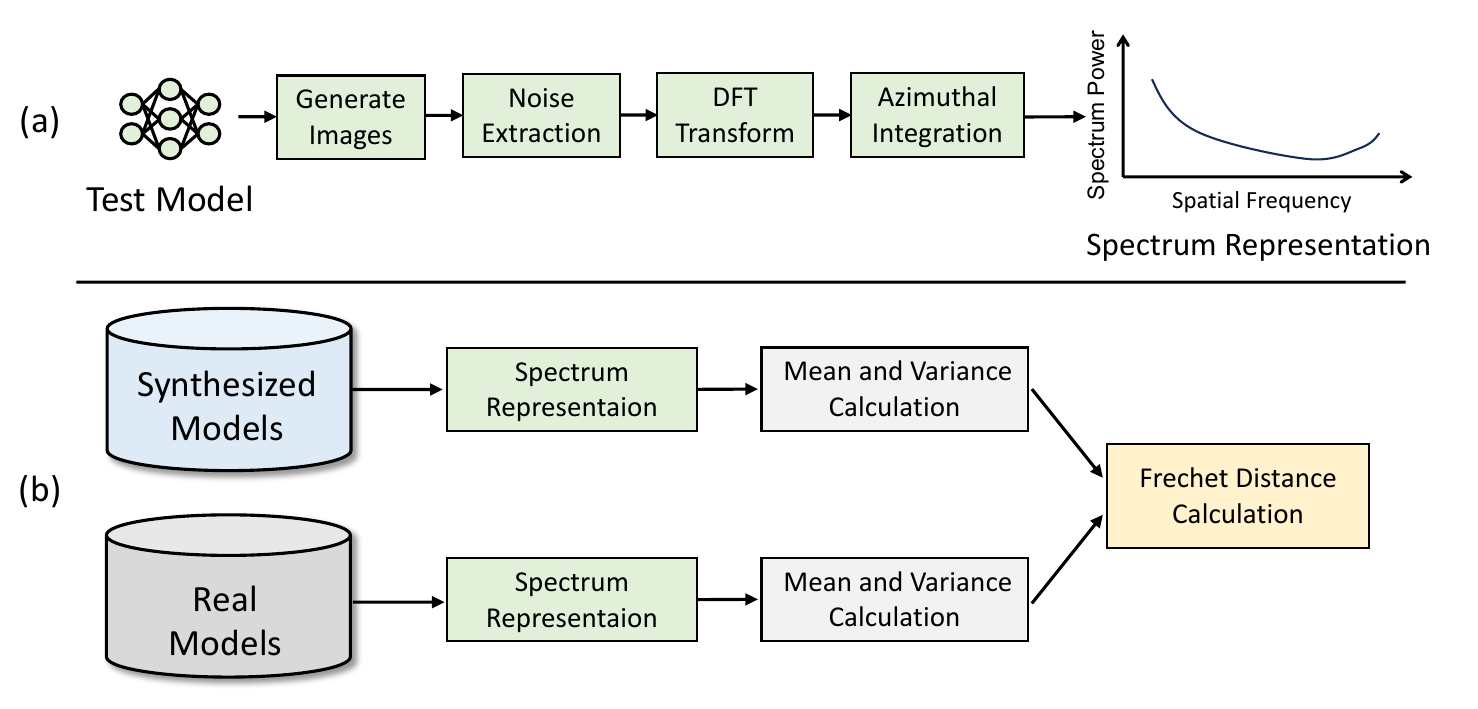}
    \caption{(a) Spectrum pattern representation on a single model. (b) The calculation process of Frechet Frequency Distance.}
    \label{fig:ffd}
\end{figure}

\textbf{Fidelity.} Currently, there is no off-the-shelf tool to quantitatively measure the gap between synthetic and real models in fingerprint distributions. To address this gap, we introduce a new metric called Frechet Frequency Distance (FFD), inspired by the Frechet Inception Distance (FID)~\cite{heusel2017gans}. The choice of Frechet Distance is made due to its practical utility and widespread acceptance for comparing distributions, especially in contexts where these the mean and covariance effectively captures differences in the average content and the variation of the data. Different from FID that is used to assess the fidelity of generated images by comparing the distribution distance of inception features, FFD assess the fidelity of the fingerprints of synthetic models by comparing the distribution distance of spectrum features. The FFD is calculated through two primary steps:
\begin{table}[tbp]
\centering
\caption{Compare the synthesis fidelity when using all synthesis models and reducing each synthesis option.}
\label{tab:fid_compare}
\begin{tabular}{cccccccccc}
\toprule[1pt]
Models & All & w/o K & w/o L & w/o S  \\  
FFD & 3.30 & 3.57 & 3.41 & 3.51 \\
\midrule
Models & w/o U  & w/o A & w/o N & w/o seed \\  
FFD & 9.81 & 4.46 & 3.02 & 3.39 \\
\bottomrule[1pt]
\end{tabular}
\end{table}
\begin{itemize}
    \item \textbf{(Step I) Spectrum Pattern Representation:} We extract the spectrum pattern of generated images to represent the fingerprint pattern of each synthesized and real models. As depicted in Figure~\ref{fig:ffd}(a), for each model, we initially generate $N$ (e.g., 100) images\footnote{We provide ablation study on the value of $N$ in Appendix~\ref{sec:appendix_ablation}.}, and then, following~\cite{corvi2023intriguing}, we enhance the spectrum patterns by applying a noise extractor~\cite{zhang2017beyond} to remove semantic contents from these images. The resultant noise images are then used to compute a reduced 1D power spectrum~\cite{durall2020watch} by Azimuthal Integration (AI) over radial frequencies on the 2D Fourier spectrum. We calculate the averaged spectrum from these $N$ images to serve as the spectrum pattern representation for each model.

    \item \textbf{(Step II) Frechet Distance Calculation:} As illustrated in Figure~\ref{fig:ffd}(b), we first compute the mean and covariance of the spectrum representations for all real-world models in Table~\ref{tab:dataset}, which covers the mainstream generative models ranging from GAN, VAE, Flow and Diffusion. Then in the same way, we calculate the mean and covariance of the spectrum representations for synthetic models in Section~\ref{subsec:model_synthesis}. The Fréchet distance is then calculated between these distributions. Specifically, for the mean $\mu_r$ and covariance $\Sigma_r$ of real distribution, and the mean $\mu_s$ and covariance $\Sigma_s$ of synthetic distribution, their distance is given by:
    \begin{align}
    \text{FFD} = \left\|\mu_r-\mu_s\right\|^2 + \operatorname{Tr}\left(\Sigma_r + \Sigma_s - 2\left(\Sigma_r \Sigma_s\right)^{1/2}\right),
    \end{align}
    where $\operatorname{Tr}$ represents the trace of a matrix. A low FFD score indicates that the synthetic models closely mimic the real models in terms of spectrum patterns.
\end{itemize}
In Table~\ref{tab:fid_compare}, we evaluate the Frechet Frequency Distance of synthetic models against real models, varying synthesis choices. The variations in the architecture include: ``w/o K" and ``w/o L," indicating the construction with a single down/up block and a single layer, respectively; ``w/o S," using only one type of sequence — normalization and activation after convolution; ``w/o U," utilizing only bilinear interpolation for upsampling; and ``w/o A" and ``w/o N," which denote the absence of activation and normalization layers, respectively. We also consider not using diverse seeds for each architecture, denoted as ``w/o seed". The results displayed in the table indicate that reducing most synthesis options tends to increase the Frechet Frequency Distance (FFD), pointing to a poorer alignment between the distributions. Notably, eliminating the diversity in activation functions and upsampling methods results in the most significant increases in FFD. This may be because these factors, when varied, mostly increase the diversity of synthetic models. 

\textbf{Efficiency.} We measured the time required to train these models. For models with single generative block ($K=1$), the averaged required training time per model is 53 seconds, while those with $K=2$ required 113 seconds each. In total, training all the synthetic models takes 133 hours on a single 3090 GPU. This is considerably more efficient than the time investment needed for training state-of-the-art generative models. For example, it takes 14 days 22 hours on single V100 GPUs to train a StyleGAN~\cite{karras2019style} model that generate with a 256$\times$256 resolution\footnote{https://github.com/NVlabs/stylegan}.

\subsection{Training the Fingerprint Extractor}
\label{subsec:fingerprint_extractor}
\begin{figure*}
\setlength{\abovecaptionskip}{0mm}
\centering
\includegraphics[width=\linewidth]{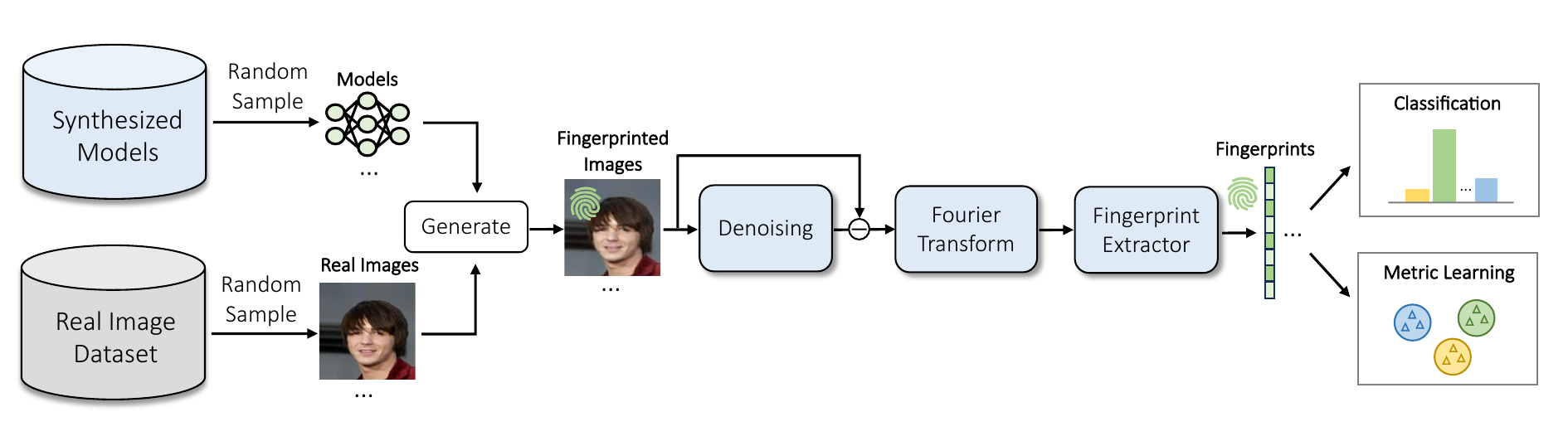} 
\caption{Training the fingerprint extractor based-on synthetic models.}
\label{fig:method}
\end{figure*}

After obtaining the synthetic models, these models are utilized to train the fingerprint extractor. The goal of this extractor is to identify unique fingerprints embedded within images generated by different models, thereby enabling differentiation among them. As illustrated in Figure~\ref{fig:method}, the training procedure involves the synthetic model pool that we construct in the above section, and a real image dataset pool.

In the training phase, a batch of real images is randomly sampled from the real image dataset pool. Simultaneously, an equivalent number of synthetic models are selected from the synthetic model pool. Each real image $I_\text{real}$ from the batch is processed through a synthetic model $M_k$, producing a generated image $I_k$ that contains distinct fingerprints characteristic of the specific synthetic model used. 
\begin{align}
    I_k = M_k (I_\text{real})
\end{align}
To reveal the underlying fingerprint patterns inherent in the images, we follow a common preprocessing technique~\cite{kirchner2009resampling, guo2023exposing} to strip away low-frequency semantic content through a denoiser by~\cite{zhang2017beyond}. The extracted noises are then transformed using a Discrete Fourier Transform to facilitate the extraction of their spectrum properties. The spectrum obtained from this transformation are fed into the fingerprint extractor, which is built upon the ResNet50 architecture. 
\begin{align}
    h_k &= f_{\operatorname{fp\_extractor}} \lp \gF (I_k^{\prime}) \rp \\
 \text{ with }   I_k^{\prime} &= I_k - f_{\operatorname{denoise}}(I_k)
\end{align}
where $f_{\operatorname{denoise}}$ and $f_{\operatorname{fp\_extractor}}$ denote the denoiser and fingerprint extractor, $\gF$ denotes the Discrete Fourier Transform, $h_k$ is the extracted fingerprint embedding.

To enhance the model's performance, we combine classification and metric loss to enhance the discrimination of learned fingerprint embedding. The classification head employs a multi-class cross-entropy (CE) loss function, which enable the fingerprint extractor to accurately classify the originating synthetic model of each input image. The projection head is followed by a triplet loss introduced in~\cite{weinberger2009distance}, which encourages generated images from the same generator to be as similar as possible in the feature space while promoting a larger dissimilarity between images from different generators. 
\begin{align}
\mathcal{L} = \mathcal{L}_{\text{CE}}(h_k, y_k) + \mathcal{L}_{\text{Triplet}}(h_a, h_p, h_n)
\end{align}
where $y_k$ is the true label for $k$-th synthetic model. $h_a$, $h_p$, and $h_n$ are the anchor, positive, and negative fingerprint embedding from a triplet set. Positive and negative mean the embedding is from the same model with the anchor embedding or not.

\section{Experiments}
\label{sec:exp}

In the experimental section, we aim to answer the following evaluation questions:\\
\textbf{EQ1} How do the synthesis, training, and inference options affect the generalization ability of the fingerprint extractor? \\
\textbf{EQ2} How does our method perform compared with existing methods in model identification and verification scenarios? \\
\textbf{EQ3} Can we apply the trained fingerprint extractor in more complex attribution tasks, such as model linage analysis? \\
\textbf{EQ4} How similar are synthetic models to real models in terms of their 2D spectrum visualizations?

\subsection{Experimental Details}
\noindent \textbf{Testing Models.} Although our analysis and synthesis strategy primarily focuses on CNN-based generative models, our evaluation includes a broader scope of models as shown in Table~\ref{tab:dataset}. In addition to CNN-based models, we also assess the generalization performance on Transformer-based models, which share many common components with CNN-based models. And for CNN-based category, we include a variety of models such as CNN, VAE, Flow, and Diffusion models. All models in the CNN-based and Transformer-based category are unconditional models, which are trained on the dataset in the ``Image Source" column with different resolutions. Furthermore, the last row of the table encompasses state-of-the-art Text2Image models such as Stable Diffusion v1.5, Stable Diffusion v2.1, Stable Diffusion XL, Glide, DalleE-mini, DalleE-2, and DalleE-3. These models are not restricted to specific domains, and diverse in architecture type, training dataset, and resolution. Although these models are primarily for multimodal text-to-image generation, they share structural similarities with GAN-based unimodal generators. For example, the foundational architecture of the Stable Diffusion v1 series employs a U-Net structure, where its upsampling blocks are also built using standard components such as normalization, activation, convolution, and upsampling layers. Furthermore, as highlighted in ~\cite{corvi2023intriguing, corvi2023detection, bammey2023synthbuster}, no generator to date is entirely artifact-free, with frequency-domain artifacts still evident even in state-of-the-art multimodal generative models.

\begin{table*}
\centering
\caption{Models highlighted in \colorbox{gray!20}{gray} are considered classical models and are used to train the existing methods being compared. The remaining models are categorized as emerging models and have not been seen by the existing methods.}
\label{tab:dataset}
\begin{tabular}{llllll}
\toprule[1.2pt]
Catogory & Type & Image Source & Res.  & Models  \\
\midrule
\multirow{5}{*}{CNN-based} & \multirow{2}{*}{GAN} & CelebA & 128 & \cellcolor{gray!20} ProGAN~\cite{karras2017progressive}, MMDGAN~\cite{binkowski2018MMD},SNGAN~\cite{zhang2019self},  StarGAN~\cite{choi2018stargan}, AttGAN~\cite{he2019attgan} \\
\cmidrule{3-5}
& & CelebAHQ/FFHQ & 1024 & \cellcolor{gray!20} ProGAN, StyleGAN~\cite{karras2019style}, StyleGAN2~\cite{karras2020stylegan2}, StyleGAN3-t\cite{karras2021alias}, StyleGAN3-r\cite{karras2021alias} \\
\cmidrule{2-5}
& VAE & CelebA & 64 & \cellcolor{gray!20} VanillaVAE~\cite{kingma2013auto}, BetaVAE~\cite{higgins2017beta}, DisentangledBetaVAE~\cite{burgess2018understanding}, InfoVAE~\cite{zhao2017infovae} \\
\cmidrule{2-5}
& Flow & CelebA & 128 & \cellcolor{gray!20} ResFlow~\cite{chen2019residual}, Glow ~\cite{kingma2018glow}\\
\cmidrule{2-5}
& Diffusion & Lsun-Bedroom & 256 & ADM~\cite{dhariwal2021diffusion}, DDPM~\cite{ho2020denoising},LDM~\cite{rombach2022high}, PDNM~\cite{liu2022pseudo} \\
\midrule
\multirow{2}{*}{Transformer-based} & \multirow{2}{*}{GAN} & FFHQ & 256 & GANsformer~\cite{hudson2021generative}, StyleSwin~\cite{zhang2022styleswin} \\
\cmidrule{3-5}
& & CelebA & 64 & TransGAN~\cite{jiang2021transgan}, Styleformer~\cite{park2022styleformer} \\
\midrule
Text2Image Models & - & -  & - & SD1.5~\cite{rombach2022high}, SD2.1, SDXL, MJ~\cite{mj}, Glide~\cite{nichol2021glide}, DalleE-mini~\cite{dalle-mini}, DalleE-2~\cite{ramesh2022hierarchical}, DalleE-3~\cite{betker2023improving}\\ 
\bottomrule[1.2pt]
\end{tabular}
\end{table*}

\begin{figure}[t]
    \centering
    \includegraphics[width=\linewidth]{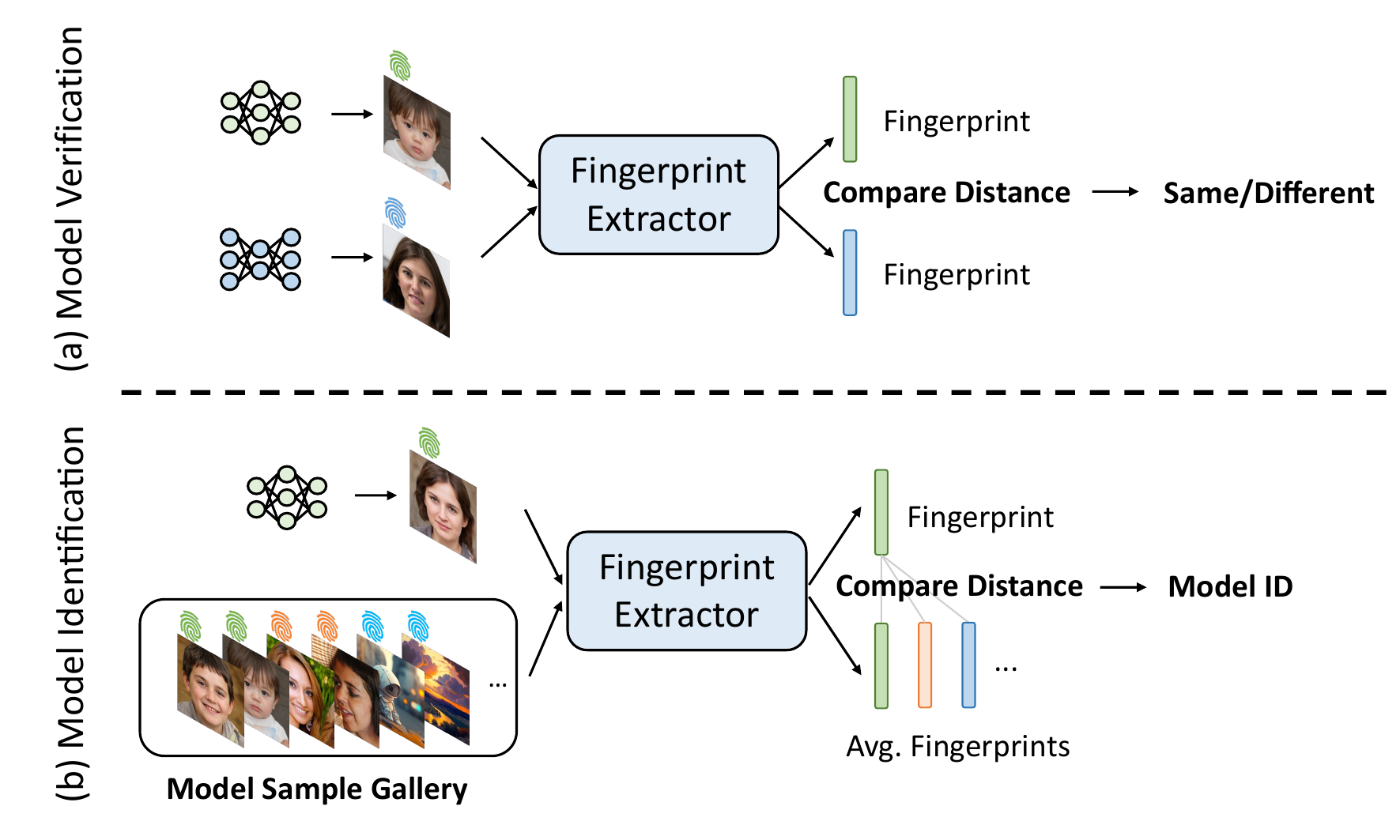}
    \caption{Illustration of the model verification (above) and identification (below) problem.}
    \label{fig:2setup}
\end{figure}
\noindent \textbf{Testing Scenarios.} Similar to the face identification and verification problems encountered in the field of face recognition, we consider utilize the fingerprint extractor in two model attribution problems, the 1:1 model verification and 1:N model identification problem, as shown in Figure~\ref{fig:2setup}. Settings and experimental setups for the two problems are illustrated below:
\begin{itemize}
    \item \textbf{Model Verification}: the 1:1 model verification problem aims to verify whether two generated images are from the same model or not. Model verification is performed by comparing the cosine similarity of the extracted fingerprint. To assess the performance, we generate 1,000 images for each model under test. Then we randomly selected 10,000 pairs out of these images, which consist of 5,000 negative pairs (images from different models) and 5,000 positive pairs (images from the same model). The evaluation employs metrics commonly used in face verification tasks: accuracy and the Area Under the Receiver Operating Characteristic Curve (AUC). Accuracy measures the correct verification of pairs as originating from the same or different models. We report the best accuracy across all similarity thresholds to assess the overall effectiveness of the verification system. The AUC provides a comprehensive measure of performance across all possible threshold levels, reflecting the trade-off between the true positive rate and the false positive rate. 
    \item \textbf{Model Identification}: the 1:N model identification problem aims to identify the specific generative model used to create a given image among N models. N is the total number of models in the model gallery. To assess the performance of model identification, we generate 1,000 images for each model under test. Out of these, 10 images are sampled to create the fingerprint for each model, while the rest serve as test samples. Each model's fingerprint is collected to form the model gallery. We compare the test sample with each fingerprint in the gallery and classify it into the class with the highest cosine similarity, which corresponds to the closest fingerprint distance. Performance is evaluated using classification accuracy and the F1 score.
\end{itemize}

\noindent \textbf{Compared methods.} To compare with existing methods, we derive two variants of our method: 1) \textit{Ours (Only-Syn)}, which is only trained on synthetic models without having seen any real-world models. 2) \textit{Ours (Fine-tune)}, which fine-tune the fingerprint extractor trained on synthetic models for one epoch using samples of models highlighted in \colorbox{gray!20}{gray} in Table~\ref{tab:dataset}.

We compare against seven existing model attribution methods. They are Marra\etal~\cite{marra2019gans}, Yu\etal~\cite{yu2019attributing}, DNA-Det~\cite{yang2022aaai}, RepMIX~\cite{bui2022repmix}, POSE~\cite{yang2023progressive}, and Abady\etal~\cite{abady2024siamese}. We train them on models in gray in Table~\ref{tab:dataset}. 
Although the works in~\cite{marra2019gans, yu2019attributing, yang2022aaai, bui2022repmix} are proposed to evaluate in a closed-set scenario, their the layers before the classification head can be adapted as a fingerprint extractor for use in a zero-shot model attribution context. This adaptation is implemented to evaluate the model identification and verification problems.

\subsection{Ablation Study (EQ1)}
\label{subsec:ablation}

In the ablation study section, we examine several key factors affecting the generalization performance of the fingerprint extractor. These factors include: the types of synthetic models and real images used for creating fingerprinted images, and the implementation of a sliding window strategy during the inference phase. This comprehensive evaluation helps in understanding how each component contributes to the effectiveness of the fingerprint extraction. Our default configurations within the tables are highlighted in \colorbox{gray!20}{gray}.
In this section, we evaluate on the \textit{Only-Syn} version of our method, without including any real generative models in training.

\begin{table}[t]
\centering
\caption{Ablation study on the architecture of synthetic models.}
\label{tab:ablation_arch}
\scalebox{1.1}{
\begin{tabular}{lcccccccccccccccccc}
\toprule[1pt] 
\multirow{2}{*}{Syn Models} & \multicolumn{2}{c}{Verification} &&\multicolumn{2}{c}{Identification} \\  
\cmidrule{2-3} \cmidrule{5-6} 
& Acc & AUC && Acc & F1 \\
\midrule
w/o K (K=1)	&87.04	&94.57	&&92.69	&92.51 \\
w/o L (L=1)	&87.54	&95.08	&&90.04	&89.86 \\
w/o S (conv-norm-act)	&89.29	&95.56	&&93.91	&93.75 \\
w/o U (bilinear)	&87.68 & 94.18 && 88.96 & 88.64 \\
w/o A (no act)	&84.75	&92.19	&&87.03	&86.80 \\
w/o N (no norm)	&87.63	&94.80	&&93.47	&93.40 \\
\rowcolor{gray!20} All  & 89.73	& 95.69 && 94.03	& 93.84 \\
\bottomrule[1pt]
\end{tabular}}
\end{table}

\begin{table}[t]
\centering
\caption{Ablation study on the number of seeds to train models based on each synthetic architecture.}
\label{tab:ablation_seed}
\scalebox{1.1}{
\begin{tabular}{lcccccccccccccccccc}
\toprule[1pt] 
\multirow{2}{*}{Seed Num} & \multicolumn{2}{c}{Verification} &&\multicolumn{2}{c}{Identification} \\  
\cmidrule{2-3} \cmidrule{5-6} 
& Acc & AUC && Acc & F1 \\
\midrule
1	&89.87	&96.11	&&91.04	&90.89 \\
10 & 88.00	&94.70 && 93.03	&92.87\\
15	&88.87	&95.37	&&93.96	&93.78 \\
\rowcolor{gray!20} 20 &89.73	&95.69	&&94.03	&93.84 \\
\bottomrule[1pt]
\end{tabular}}
\end{table}



\begin{table}[t]
\centering
\caption{Ablation study on the real dataset for fingerprinted image generation based-on synthetic models.}
\label{tab:ablation_dataset}
\scalebox{1.1}{
\begin{tabular}{lccccccccc}
\toprule[1pt]
\multirow{2}{*}{Real Dataset} & \multicolumn{2}{c}{Verification} &&\multicolumn{2}{c}{Identification} \\  
\cmidrule{2-3} \cmidrule{5-6} 
& Acc & AUC && Acc & F1 \\
\midrule
CelebA& 87.50	&94.60& & 	93.90& 	93.78\\
LSUN  & 89.72	&95.53& &	93.62& 	93.48 \\
\rowcolor{gray!20} CelebA + LSUN  & 89.73	&95.69& & 	94.03& 	93.84 \\
\bottomrule[1pt]
\end{tabular}}
\end{table}

\begin{table}[t!]
\centering
\caption{Ablation study on the sliding window.}
\label{tab:ablation_sw}
\scalebox{1.1}{
\begin{tabular}{lccccccccc}
\toprule[1pt]
\multirow{2}{*}{Slide Strategy} & \multicolumn{2}{c}{Verification} &&\multicolumn{2}{c}{Identification} \\  
\cmidrule{2-3} \cmidrule{5-6} 
& Acc & AUC && Acc & F1 \\
\midrule
w/o slide window & 88.42	&94.63	&&87.92	&87.68 \\
2$\times$2 slide window &  89.88	&95.38 && 91.84&	91.79 \\
\rowcolor{gray!20} 3$\times$3 slide window &89.73	& 95.69  &&94.03	&93.84 \\
4$\times$4 slide window & 89.68	&95.60	&& 94.03	&93.81	\\
\bottomrule[1pt]
\end{tabular}}
\end{table}

\begin{figure}[t]
    \centering
    \includegraphics[width=\linewidth]{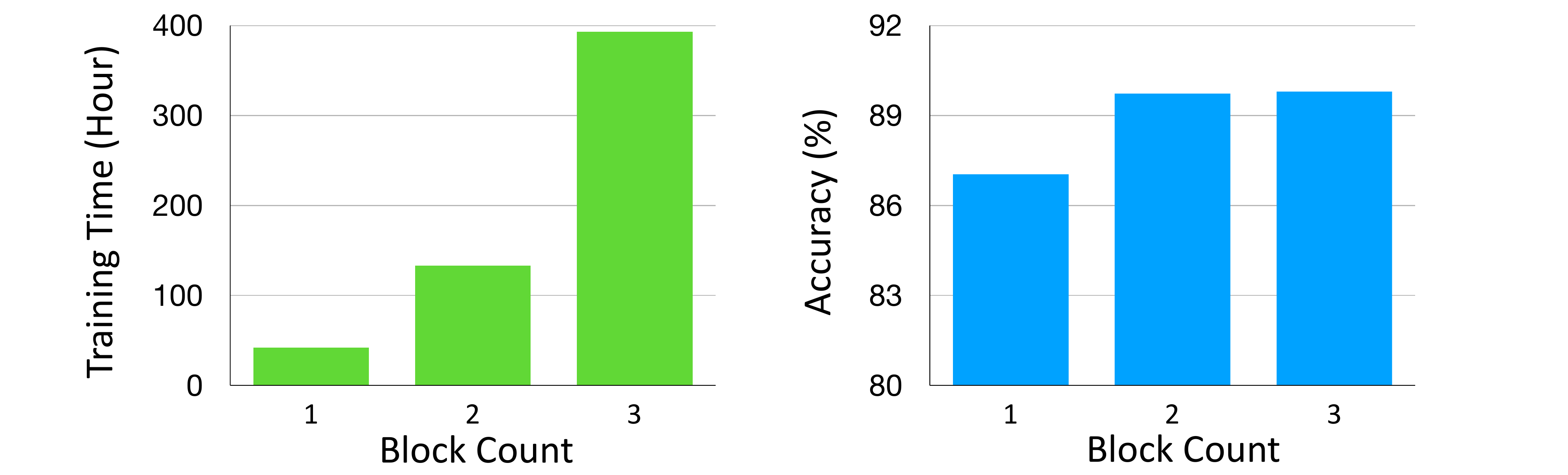}
    \caption{(Left) Totaling training time for synthetic models with a maximum generative block count of 1, 2, and 3, corresponding to 2880, 5760, and 8640 models, respectively. (Right) The model verification accuracy by the fingerprint extractor trained on synthetic models with a maximum block count from 1 to 3.}
    \label{fig:block_num_ablation}
\end{figure}

\noindent \textbf{Synthetic Options.} 
To evaluate the impact of individual synthesis options of synthetic models on the generalization of the trained fingerprint extractor in model verification and identification tasks, we conducted an ablation study. The results, presented in Table~\ref{tab:ablation_arch}, demonstrate that omitting any synthesis option reduces the generalization capabilities in model identification and verification tasks. Notably, the removal of the upsampling and activation layers significantly affects generalization performance. This observation aligns with the Frechet Frequency Distance scores listed in Table~\ref{tab:fid_compare}, where the absence of activation and upsampling variation results in the highest scores, indicating the lowest fidelity of synthetic models. 

\noindent \textbf{Depth of Synthetic Models.} We also discuss how depth of synthetic models influences the attribution performance. In Figure~\ref{fig:block_num_ablation}, we plot the total training time required for synthetic models with a maximum block count ranging from 1 to 3. For models with a single generative block, the averaged training time for all models is 42 GPU hours, involving models with 2 generative blocks, the training time is 133 hours, and increase to 393 hours when adding models with 3 generative blocks. Meanwhile, we observe an improvement in model verification accuracy when increasing the block count to 2, but no significant performance increase with 3 blocks. This indicates that using synthetic models with a maximum block count of 2 achieves the best trade-off between synthesis computation costs and model attribution performance.

\begin{table}[t]
\centering
\caption{Performance comparison of model identification.}
\label{tab:1vn_result}
\scalebox{1.1}{
\begin{tabular}{lccccccccccc}
\toprule[1pt]
\multirow{2}{*}{Method} & \multicolumn{2}{c}{Classical Models} & & \multicolumn{2}{c}{Emerging Models} \\ 
\cmidrule{2-3} \cmidrule{5-6} 
& Acc  & F1  & &  Acc  & F1 \\
\midrule
Marra et al.~\cite{marra2019gans} &61.00	&55.01	&&26.72	&22.33 \\
Yu et al.~\cite{yu2019attributing} & 96.28	&96.27	&&31.93	&30.67 \\
DNA-Det~\cite{yang2022aaai} &  98.55	&98.55	&&47.41	&46.60 \\
RepMix~\cite{bui2022repmix} &  94.34 & 94.29 && 26.63 & 25.35 \\
Abady et al.~\cite{abady2024siamese} &  97.12 & 97.09	&& 48.23 & 48.15 \\
POSE~\cite{yang2023progressive} & \underline{98.93}	&\underline{98.93}	&&49.80	&48.87 \\
\midrule
Ours (Only-Syn) & 96.71	&96.70	&&\underline{93.20}	&\underline{93.11} \\
Ours (Fine-tune) & \textbf{99.88}	&\textbf{99.89}	&&\textbf{94.02}	&\textbf{93.97} \\
\bottomrule[1pt]
\end{tabular}}
\end{table}

\noindent \textbf{Real Dataset.} In Table~\ref{tab:ablation_dataset}, we evaluate how the choice of real datasets used to generate fingerprinted images with synthetic models affects the generalization performance of the trained fingerprint extractor. We consider three options for constructing the real image pool: the CelebA dataset, which exclusively contains face images; the LSUN dataset, featuring 20 different types of semantic images; and a composite of both datasets. The results indicate that the combination of the CelebA and LSUN datasets, which offers the richest semantic diversity, yields the best generalization performance. This phenomenon may be because the increasing the semantic diversity of fingerprinted images helps the fingerprint extractor focus on semantic-agnostic fingerprint features.


\noindent \textbf{Sliding Window.}
During the inference phase for extracting fingerprints from test images of varying resolutions, we implement a sliding window strategy to accommodate different input sizes. The training resolution for the fingerprint extractor is set at 128$\times$128 pixels. For testing images that exceed this resolution, we employ an equally spaced sliding window technique to divide the image into N$\times$N patches, each with 128$\times$128 pixels. These patches are then individually processed to extract fingerprints. The final fingerprint for the entire input image is obtained by averaging the fingerprints from all these patches, ensuring comprehensive coverage and utilization of the image's information for accurate fingerprint extraction. For images with resolutions smaller than 128$\times$128, we resize them to the required 128$\times$128 resolution. As indicated in Table~\ref{tab:ablation_sw}, the sliding window strategy significantly enhances model identification performance. This improvement is attributed to the strategy's ability to capture a richer set of fingerprint information within the test image. Given that the performance of $3\times3$ and $4\times4$ sliding configurations shows similar results, we use the $3\times$3 configuration for inference efficiency.

\subsection{Comparison with Existing Methods (EQ2)}
\label{subsec:compare_existing}
\noindent \textbf{Model Identification}
In Table~\ref{tab:1vn_result}, we evaluate and compare our methods—Ours (Only-Syn) and Ours (Fine-tune)—against existing methods in the 1:N model identification scenario. The results show that most existing methods achieve high identification accuracy on classical models, which are seen in their training. However, their performance significantly declines when tested on unseen emerging models, dropping below 50\%. In contrast, our method ``Ours (Only-Syn)" demonstrates strong generalization capabilities. Despite being trained only on synthesized models and not having exposure to real generative models, it achieves over 90\% identification accuracy on the two sets of models. To align more closely with existing methods, we introduce ``Ours (Fine-tune)," which is further fine-tuned using classical model images. This adjustment improves the identification accuracy to 99\% on classical models, with a modest enhancement on emerging models. These results underscore that our fingerprint extractor, by leveraging synthetic models, exhibits superior generalization ability across real-world generative models due to its exposure to a broader model fingerprint variations. 


\begin{table}[t]
\centering
\caption{Performance comparison of model verification.}
\label{tab:1v1_verify}
\scalebox{1.1}{
\begin{tabular}{lccccccccccc}
\toprule[1pt]
\multirow{2}{*}{Method} & \multicolumn{2}{c}{Classical Models} & & \multicolumn{2}{c}{Emerging Models} \\ 
\cmidrule{2-3} \cmidrule{5-6} 
& Acc  & AUC  & &  Acc  & AUC \\
\midrule
Marra et al.~\cite{marra2019gans} &66.10	&68.99	&&59.37	&59.46\\
Yu et al.~\cite{yu2019attributing} &  96.80	&99.27	&&63.84	&68.97\\
DNA-Det~\cite{yang2022aaai} &  \textbf{99.31} &\textbf{99.94}	&&69.64	&77.54\\
RepMix~\cite{bui2022repmix} &  99.15 & 97.44 && 61.09  & 57.92 \\
Abady et al.~\cite{abady2024siamese} &  95.10	& 96.98	&& 64.52 &68.98\\
POSE~\cite{yang2023progressive} &  98.76	&99.91	&&66.94	&72.85\\
\midrule
Ours (Only-Syn) & 92.01	&96.90	&&\textbf{85.88}	&\textbf{92.66}\\
Ours (Fine-tune) &  \underline{99.21} & \underline{99.93}	&&\underline{85.62} & \underline{92.55}\\
\bottomrule[1pt]
\end{tabular}}
\end{table}

\noindent \textbf{Model Verification.} In Table~\ref{tab:1v1_verify}, we compare our methods against existing methods in the 1:1 model verification scenario, which involves determining whether two input images are generated by the same model. Similar to the findings in the 1:N model identification scenario, existing methods exhibit high verification accuracy for seen models but show significant performance degradation when applied to unseen models. Our method, Ours (Only-Syn), trained solely on synthetic models, demonstrates balanced performance across the two groups of model. This indicates a robust ability to generalize from synthetic model to real-world generative models. When we fine-tune the fingerprint extractor with images from classical models, as in Ours (Fine-tune), the performance on classical models is up to par with existing methods while maintaining superior performance on unseen emerging models. 

\begin{table}[t]
\centering
\caption{Silhouette Score (SS) and Davies-Bouldin Index (DBI) comparison for fingerprint representations (higher values of Silhouette Score and lower values of Davies-Bouldin Index indicate better performance).}
\label{tab:cluster_result}
\scalebox{1.1}{
\begin{tabular}{lccccccccccc}
\toprule[1pt]
\multirow{2}{*}{Method} & \multicolumn{2}{c}{Classical Models} & & \multicolumn{2}{c}{Emerging Models} \\ 
\cmidrule{2-3} \cmidrule{5-6} 
& SS $\uparrow$  & DBI $\downarrow$ & &  SS$\uparrow$  & DBI $\downarrow$ \\
\midrule
Marra et al.~\cite{marra2019gans} & -0.07 & 23.18 &&  -0.21 & 34.17\\
Yu et al.~\cite{yu2019attributing} & 0.36 & 1.21 &&  -0.05 & 12.73\\
DNA-Det~\cite{yang2022aaai} & 0.40 & 1.08 &&  0.01 &4.53 \\
RepMix~\cite{bui2022repmix} & 0.28 & 2.59 &&  -0.10 &8.48 \\
Abady et al.~\cite{abady2024siamese} & 0.38 & 1.18 &&  -0.02 &10.96\\
POSE~\cite{yang2023progressive} & \underline{0.44} & \underline{0.90} && -0.02 &3.64 \\
\midrule
Ours (Only-Syn) & 0.19 & 2.02 && \textbf{0.19} &\textbf{1.84} \\
Ours (Fine-tune) & \textbf{0.45} & \textbf{0.84} && \underline{0.18} & \underline{1.88} \\
\bottomrule[1pt] 
\end{tabular}}
\end{table}

\noindent \textbf{Evaluation of Fingerprint Representation.} In Table~\ref{tab:cluster_result}, we include additional evaluation metrics to assess the quality of fingerprint representation: 1)\textit{Silhouette Score (SS)}~\cite{rousseeuw1987silhouettes}, which measures how similar a point is to its own cluster compared to other clusters. Higher values indicate better separation between fingerprints of different models. 2) \textit{Davies-Bouldin Index (DBI)}~\cite{davies1979cluster}, which measures the ratio of intra-cluster distance to inter-cluster distance. Lower values indicate more compact and separated fingerprint representation. As shown in the table, existing methods such as DNA-Det and POSE achieve relatively high SS and low DBI on seen classical models. However, their generalization to unseen emerging models is limited, with SS decreasing and DBI increasing significantly. In contrast, our method, Ours (Only-Syn), demonstrates more consistent performance across both classical and emerging models. Notably, after fine-tuning on classical models, Ours (Fine-tune) achieves the highest SS and lowest DBI compared with existing methods, indicating superior fingerprint separation and robust generalization across diverse generative models.

\subsection{Model Lineage Analysis (EQ3)}
\begin{figure}[t]
\centering
\includegraphics[width=1\linewidth]{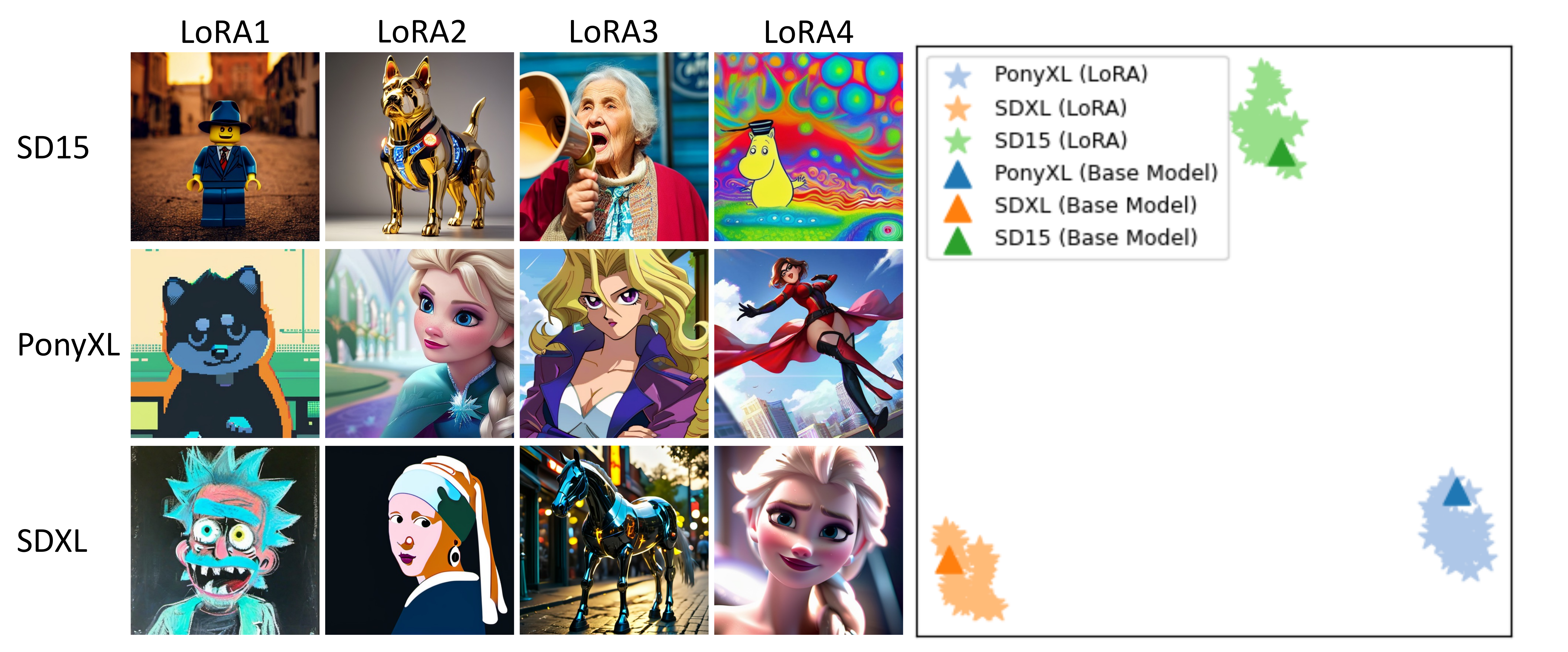} 
\begin{minipage}{0.6\linewidth}
    \centering
    \figtext{(a)}
\end{minipage}
\begin{minipage}{0.38\linewidth}
    \centering
    \figtext{(b)}
\end{minipage}
\caption{(a) Images Generated by Different LoRA-Equipped Models: 
We collect four distinct LoRA weights for each of the three Base Models: SD15, PonyXL, and SDXL. (b) t-SNE visualization of the fingerprint embeddings of the three Base Models, alongside generated images by their corresponding LoRA-equipped variants. }
\label{fig:cvitai_lora}
\end{figure}

At times, models may be stolen and utilized as the base model for training variants by other entities. Our model fingerprint extractor (Only-Syn) demonstrates the potential to trace back to the base model of such variants. To illustrate this capability, we gathered models from Civitai\footnote{https://civitai.com/}, a community renowned for its vast repository of high-quality Stable Diffusion models. The majority of these models are LoRA-variant models, fine-tuned from a base model using LoRA (Low-Rank Adaption)~\cite{hu2022lora}. We collected four distinct LoRA weights for each of the three most commonly used base models: SD15, PonyXL v6, and SDXL 1.0. Subsequently, we generated 10 images for each LoRA-equipped model and base model. Figure~\ref{fig:cvitai_lora}(a) show the generated images by these LoRA variants, which exhibit discernible differences in styles and semantics. We employing the averaged fingerprint of generated images as the fingerprint for the base model. As shown in Figure~\ref{fig:cvitai_lora}(b), the LoRA-equipped variants exhibit similar fingerprint embeddings with their respective base models. This result shows that we could easily attribute the LoRA variant to its base model comparing the extracting fingerprints from their generated images.

\subsection{Visualization Results (EQ4)}
\newcommand\subfigurewidth{0.48}

\begin{figure}
\centering
\begin{minipage}{0.24\linewidth}
    \centering
    \figtext{\textbf{Real Models}}
\end{minipage}
\begin{minipage}{0.24\linewidth}
    \centering
    \figtext{\textbf{Synthetic Models}}
\end{minipage}
\begin{minipage}{0.24\linewidth}
    \centering
    \figtext{\textbf{Real Models}}
\end{minipage}
\begin{minipage}{0.24\linewidth}
    \centering
    \figtext{\textbf{Synthetic Models}}
\end{minipage}

\begin{minipage}{\subfigurewidth\linewidth}
    \includegraphics[width=\linewidth]{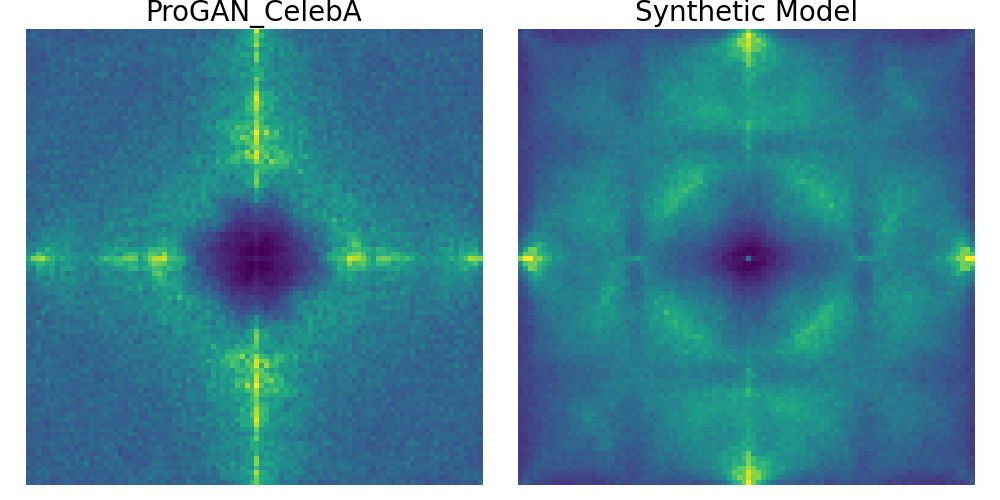}
\end{minipage}
\begin{minipage}{\subfigurewidth\linewidth}
    \includegraphics[width=\linewidth]{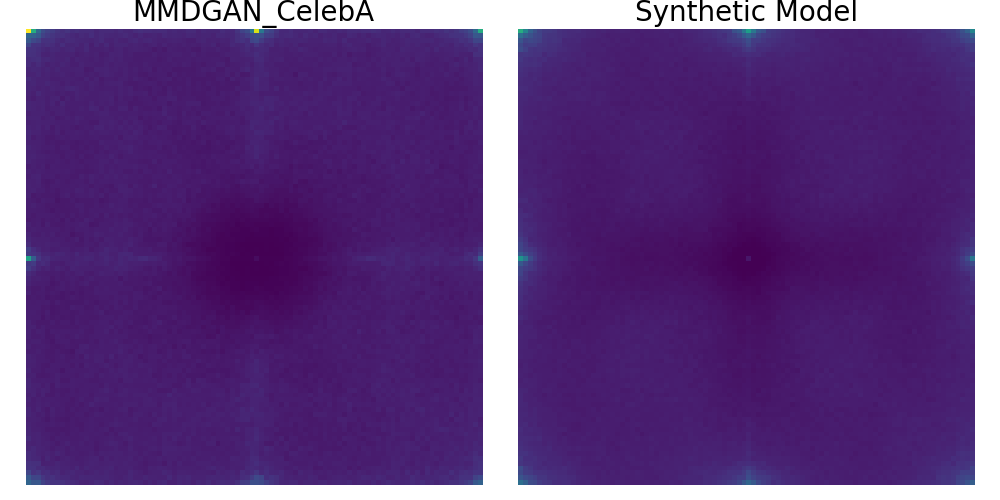}
\end{minipage}
\begin{minipage}{\subfigurewidth\linewidth}
    \includegraphics[width=\linewidth]{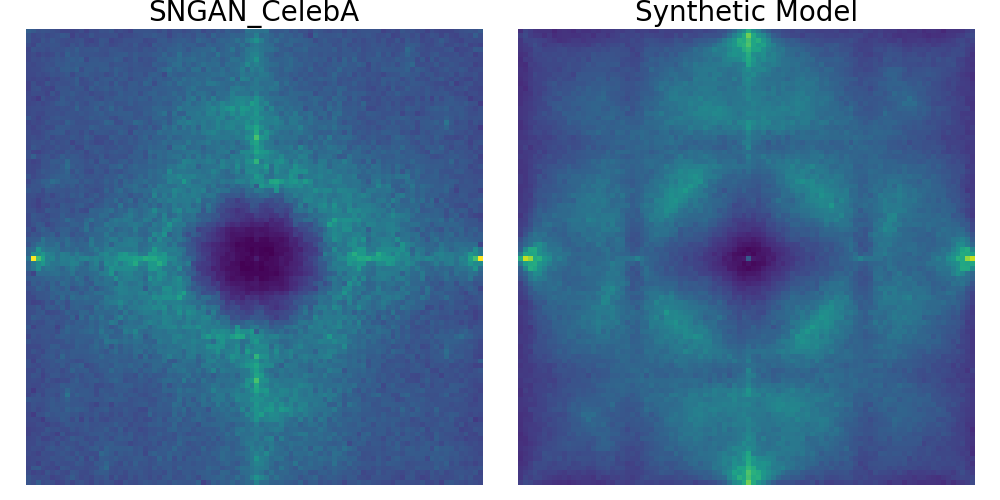}
\end{minipage}
\begin{minipage}{\subfigurewidth\linewidth}
    \includegraphics[width=\linewidth]{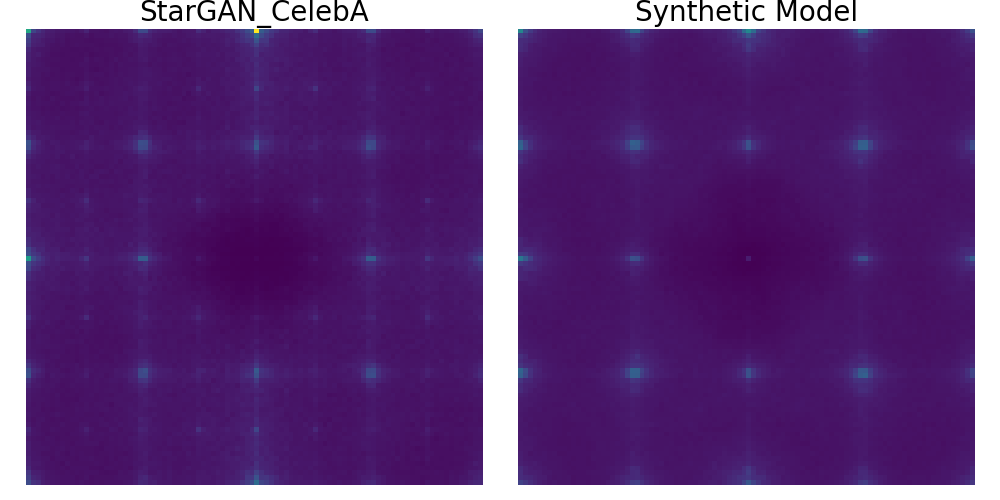}
\end{minipage}

\begin{minipage}{\subfigurewidth\linewidth}
    \includegraphics[width=\linewidth]{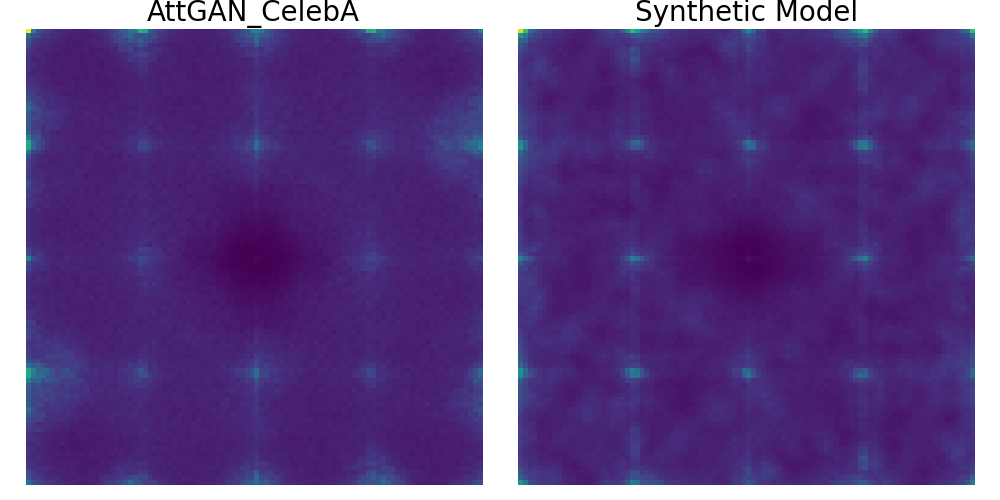}
\end{minipage}
\begin{minipage}{\subfigurewidth\linewidth}
    \includegraphics[width=\linewidth]{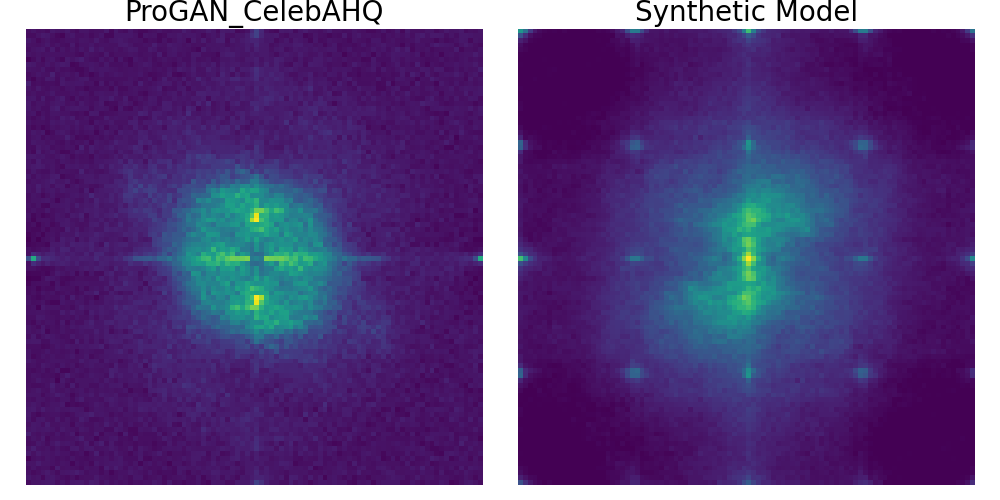}
\end{minipage}
\begin{minipage}{\subfigurewidth\linewidth}
    \includegraphics[width=\linewidth]{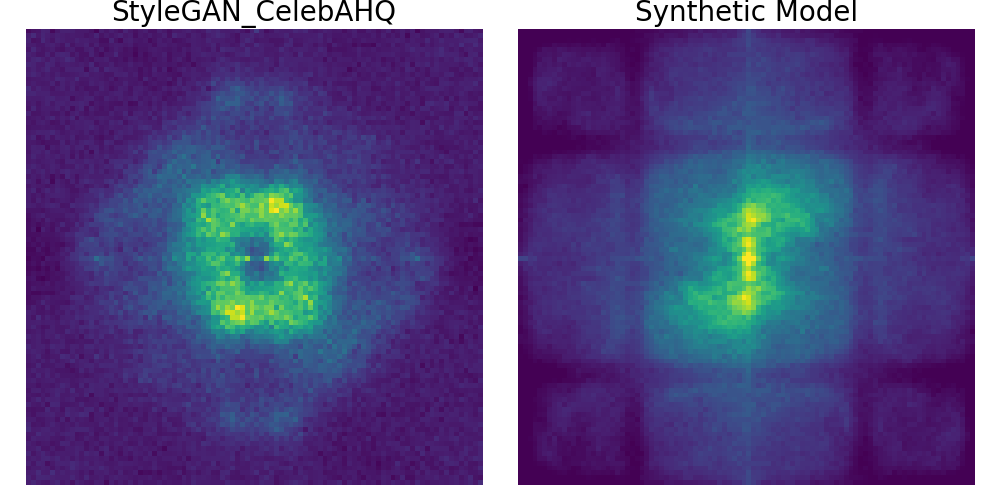}
\end{minipage}
\begin{minipage}{\subfigurewidth\linewidth}
    \includegraphics[width=\linewidth]{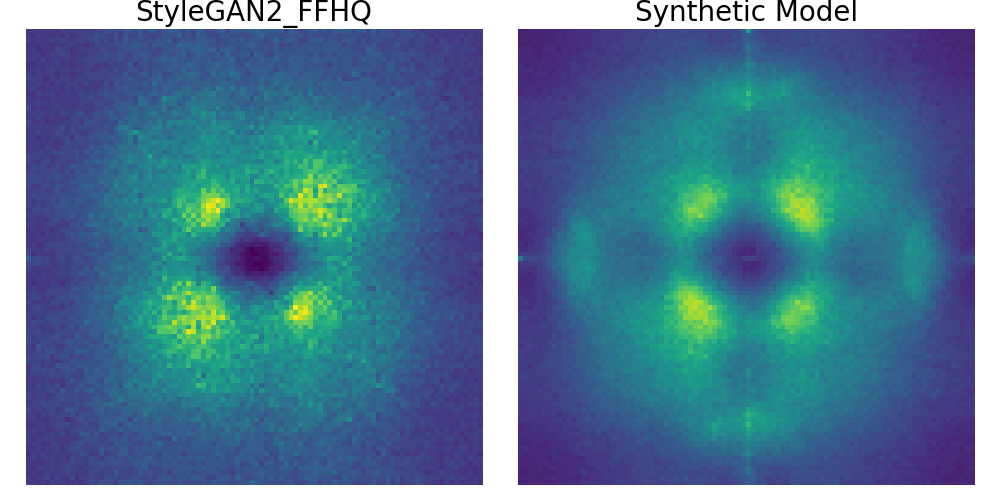}
\end{minipage}

\caption{Spectrum of real-world models and their closest synthetic models within the 5760 synthetic models.}
\label{fig:syn_spectrum_compare_part}
\end{figure}

We provide spectrum visualizations of real models and their closest synthetic models in Figure~\ref{fig:syn_spectrum_compare_part}. As shown, while the spectra of synthetic models do not exactly replicate those of real-world models, they exhibit notable similarities and diversity in frequency artifacts. This demonstrates the fidelity of our synthetic model generation process and its ability to approximate the spectral characteristics of real-world models effectively.

\section{Conclusions}

In this paper, we propose to tackle the model attribution problem in a more generalized scenario, to expand the attribution target to unseen models that are not included in training. To solve the problem, we propose a novel method by training on a large and diverse set of synthetic models, which mimic the fingerprint patterns of real generative models. The synthesis strategy is based a detailed analysis of how different model architectures and parameters influence fingerprint patterns, and is validated using two designed metrics to assess their fidelity and diversity. Experimental results demonstrate that our fingerprint extractor, trained on synthetic models, exhibits superior generalization capabilities in identifying and verifying unseen real generative models across various scenarios.

\bibliographystyle{IEEEtran}
\bibliography{reference.bib}

\clearpage

\appendix

\subsection{Additional Analysis Results of Model Fingerprint}
\label{sec:appendix_analysis}

To validate the theoretical insights in the main text, we also provide additional experiments using ProGAN, SNGAN, and StyleGAN2 models with varying types of network components below, analyzing their spectrum influence on output images.

\subsubsection{\textbf{Upsampling Effects}} 
\begin{figure}[t]
\setlength{\abovecaptionskip}{2mm}
\centering

\begin{minipage}{0.48\linewidth}
    \centering
    \figtext{SNGAN}
\end{minipage}
\begin{minipage}{0.48\linewidth}
    \centering
    \figtext{ProGAN}
\end{minipage}

\begin{minipage}{0.48\linewidth}
\includegraphics[width=\linewidth]{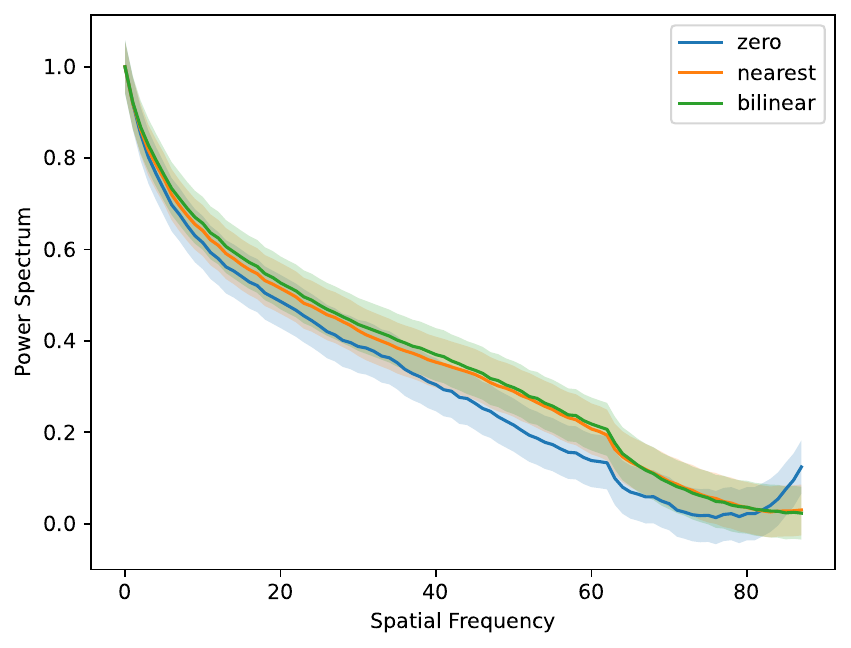}
\end{minipage}
\begin{minipage}{0.48\linewidth}
\includegraphics[width=\linewidth]{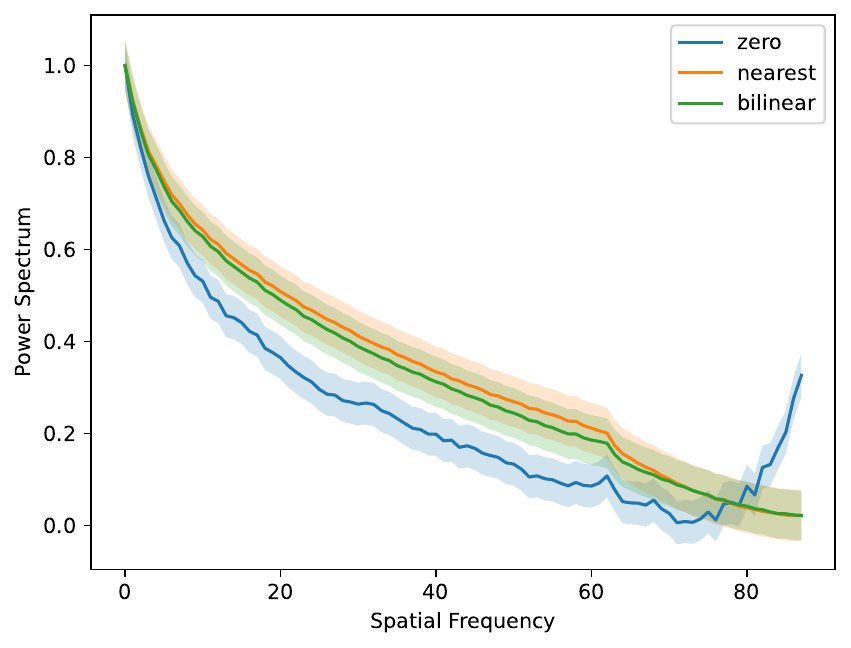}
\end{minipage}

\caption{Spectral effects of different types of upsampling layers.} 
\label{fig:up_dft}
\end{figure}

As discussed in the main text, different types of upsampling tend to leave different spectral characteristics due to differences in their upsampling kernels' shapes. To illustrate this, we trained ProGAN~\cite{karras2017progressive} and SNGAN~\cite{miyato2018spectral} models on the CelebA dataset, each with three types of upsampling layers. Figure~\ref{fig:up_dft} presents the reduced 1D power spectrum of these models, where the x-axis indicates different frequencies from low to high, and the y-axis denotes the power of these frequency components. The figure illustrates that nearest and bilinear, which act as low-pass filters, significantly reduce high-frequency content, resulting in lower power at higher frequencies on the spectrum's right side. In contrast, transposed convolution tends to preserve more high-frequency components. This is because the learned convolution kernels in transposed convolution layers are not guaranteed to be low-pass, allowing more high-frequency artifacts to remain in the generated images~\cite{zhang2019detecting}. These results are consistent with our theoretical analysis presented in the main text.

\subsubsection{\textbf{Convolution Effects}} 
\begin{figure}[t]
\centering


\begin{minipage}{0.98\linewidth}
\includegraphics[width=\linewidth]{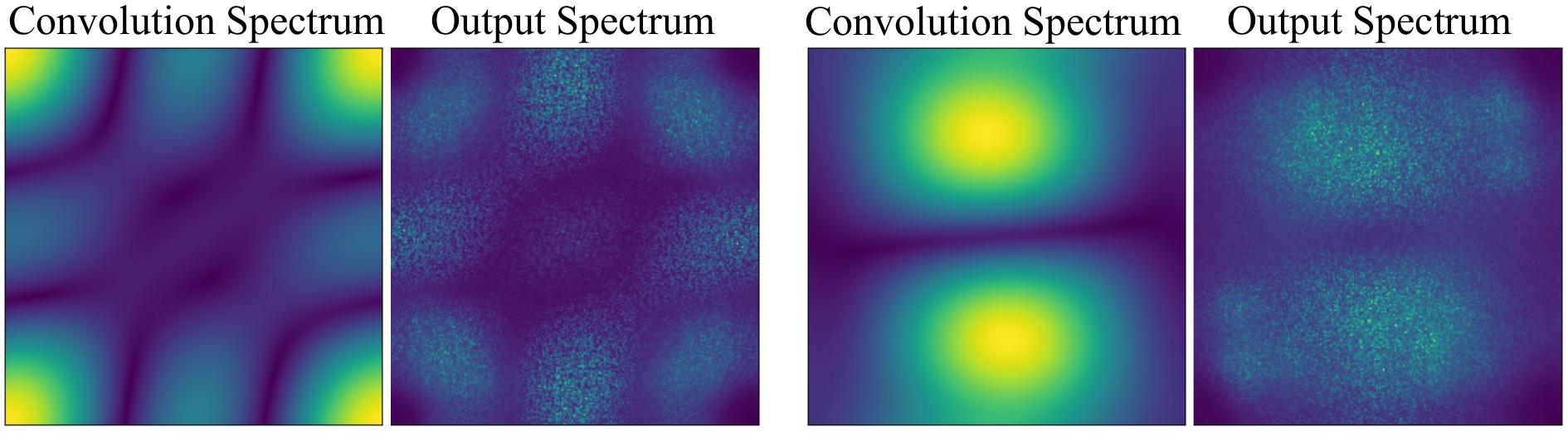}
\end{minipage}
\begin{minipage}{0.48\linewidth}
    \centering
    \figtext{(a) StyleGAN2 (FFHQ)}
\end{minipage}
\begin{minipage}{0.48\linewidth}
    \centering
    \figtext{(b) StyleGAN2 (FFHQu)}
\end{minipage}
\caption{Spectral effects of convolution parameters. This figure compares the summed spectrum of the convolution parameters with the spectrum of the output feature map.} 
\label{fig:conv_dft}
\end{figure}

As discussed in the main text, the spectral fingerprints of convolutional parameters are shaped by the spectrum characteristics of their kernels. Different kernels can lead to different emphases on certain frequency bands, thus altering the model's overall spectrum pattern. This phenomenon is illustrated in Figure~\ref{fig:conv_dft}, which shows the spectrum of two StyleGAN2~\cite{karras2020stylegan2} models that have identical architecture but differ in their parameters, trained respectively on the FFHQ and FFHQu datasets. For this analysis, we extracted the parameters from a specific output channel in the last convolutional layer of each model's final generative block. We then plot two figures for each model: the summed spectrum of the convolution kernels (left), and the spectrum of the output feature map for this channel (right). As demonstrated in the figure, for both models, the spectrum of the output feature map closely mirrors the spectrum of the convolution parameters. This alignment highlights how the convolutional layer's kernel characteristics directly shape the spectrum pattern of the output, which validates the theoretical analysis outlined in the main text. 

\subsubsection{\textbf{Nonlinear Activation Effects}} 
\begin{figure}[t!]
\setlength{\abovecaptionskip}{2mm}
\centering

\begin{minipage}{0.48\linewidth}
    \centering
    \figtext{SNGAN}
\end{minipage}
\begin{minipage}{0.48\linewidth}
    \centering
    \figtext{ProGAN}
\end{minipage}

\begin{minipage}{0.48\linewidth}
\includegraphics[width=\linewidth]{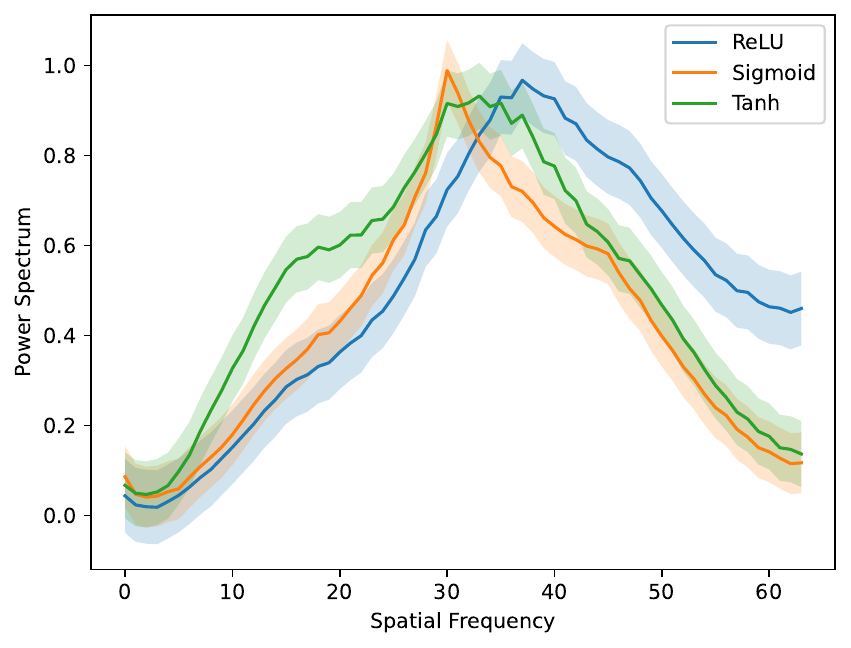}
\end{minipage}
\begin{minipage}{0.48\linewidth}
\includegraphics[width=\linewidth]{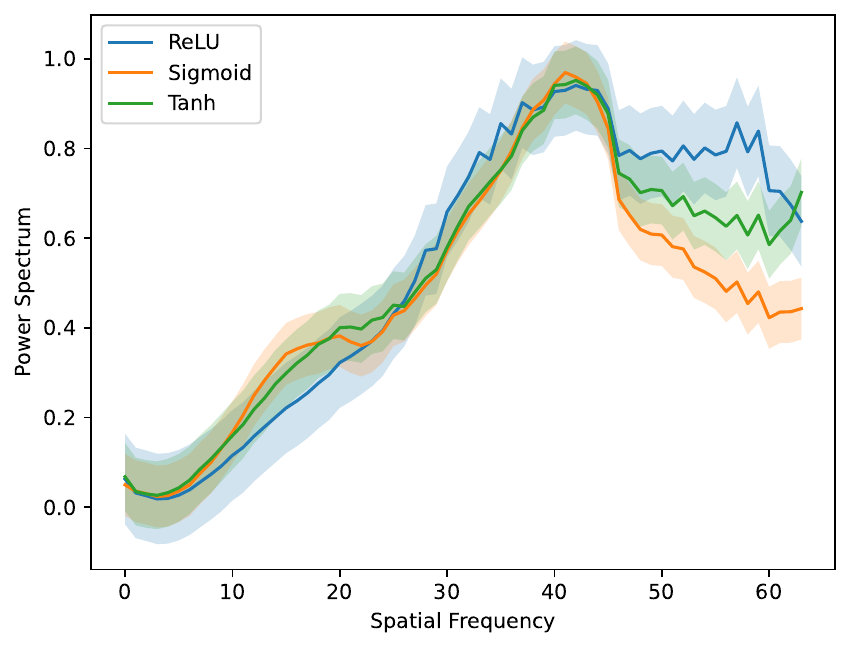}
\end{minipage}

\caption{Spectral effects of different types of activation functions.} 
\label{fig:act_dft}
\end{figure}

As discussed in the main text, the choice of activation functions can also influence a model’s frequency response, with different functions generating distinct harmonic patterns in the frequency spectrum. To verify this, we trained ProGAN and SNGAN models on the CelebA dataset, each with three different activation functions. As shown in Figure~\ref{fig:act_dft}, generative models with ReLU activation generate images with more high-frequency components than Sigmoid and Tanh, as indicated by higher values on the right side of the reduced spectrum\footnote{To make the high-frequency discrepancy more evident, we first use a denoising filter by~\cite{zhang2017beyond} with noise parameter $\sigma=1$ to remove the low-frequency semantic contents.}.

\newcommand\convsubfigurewidth{0.24}

\begin{figure}[t!]
\centering


\begin{minipage}{0.75\linewidth}
\includegraphics[width=\linewidth]{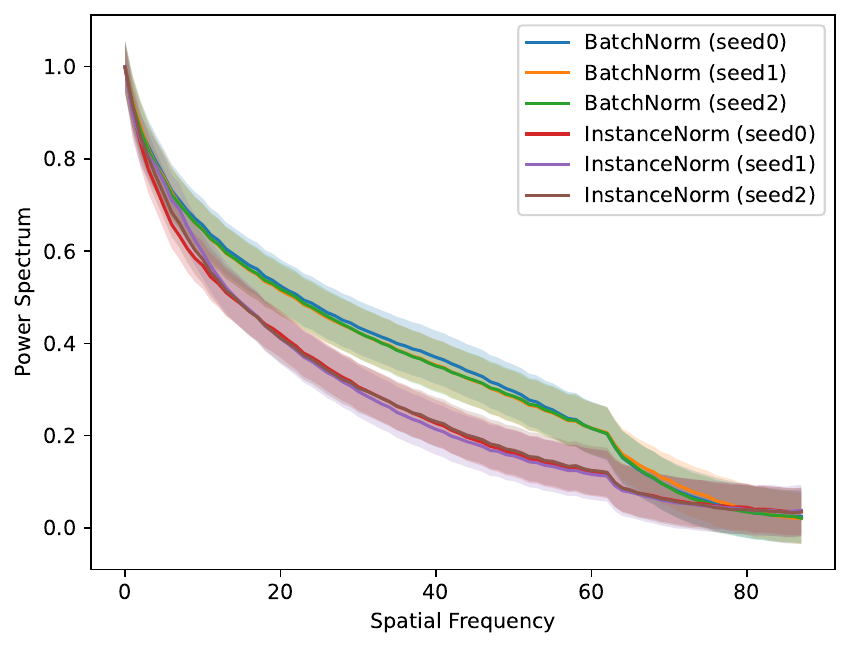}
\end{minipage}


\caption{Spectral effects of different types of normalization functions.} 
\label{fig:norm_dft}
\end{figure}

\subsubsection{\textbf{Normalization Effects}} As discussed in the main text, different normalization methods share similar mathematical formulations in the frequency domain, but they can still influence a network’s behavior in distinct ways. Variations in the optimization landscapes introduced by different normalization techniques can result in distinctive parameter distributions and, consequently, unique spectral patterns. To illustrate this, we train six SNGAN models using two types of normalization, each type with three different initial seeds. As demonstrated in Figure~\ref{fig:norm_dft}, models with the same normalization type often exhibit more consistent spectral patterns compared to those trained using different normalization methods. This underscores the impact of normalization choices on the diversity of spectrum patterns of generative models.

\subsection{Attribution Robustness Analysis}
\begin{figure}
\centering
\begin{minipage}{\linewidth}
\includegraphics[width=0.96\linewidth]{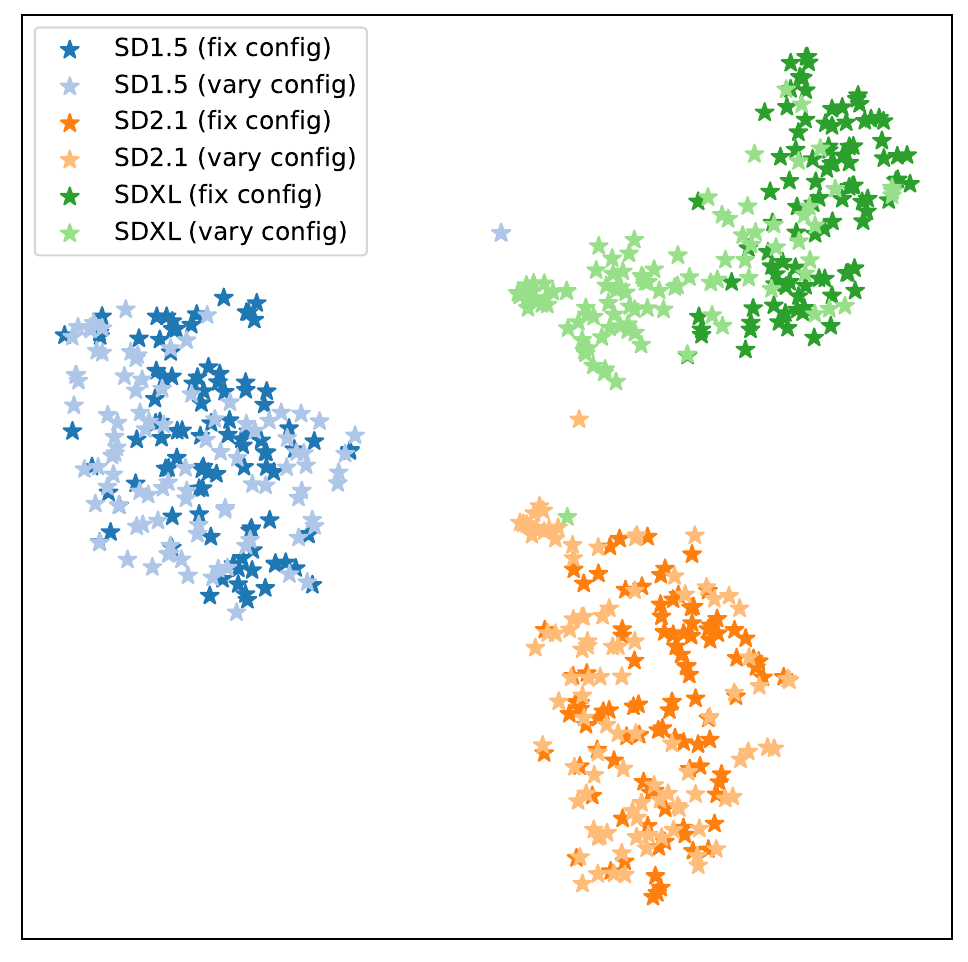}
\end{minipage}
\caption{t-SNE visualization of extracted fingerprints from diffusion images with fixed and varied generation configurations.} 
\label{fig:vary_configs}
\end{figure}

\noindent \textbf{Different Generation Configurations.} In the generation phase, diffusion models offer various configurable parameters, such as the sampling step, sampling scheduler, guidance scale, and the resolution of the generated images. These variations in setup can influence the quality of the generated images. To assess whether the extracted fingerprint remains stable amidst these variations, we conducted evaluations on three Stable Diffusion models: SD1.5, SD2.1, and SDXL 1.0. We systematically varied these configurations during image generation. Samplers from DDIM~\cite{song2020denoising}, PNDM, LMS, and Euler~\cite{karras2022elucidating} were used randomly. The guidance scale was uniformly sampled between 5 and 10, while image dimensions varied with widths and heights ranging from 256 to 1024 pixels, and sampling steps from 20 to 50. 
Figure~\ref{fig:vary_configs} illustrates the t-SNE plots of fingerprint embeddings extracted by our fingerprint extractor (Only-Syn), from models under both fixed and varied generation configurations. The results demonstrate that images with varied generation settings exhibit fingerprints as stable as those with fixed settings, underscoring the robustness of our trained fingerprint extractor to variations in diffusion generation configurations.

\begin{figure}
\centering
\begin{minipage}{\linewidth}
\includegraphics[width=0.96\linewidth]{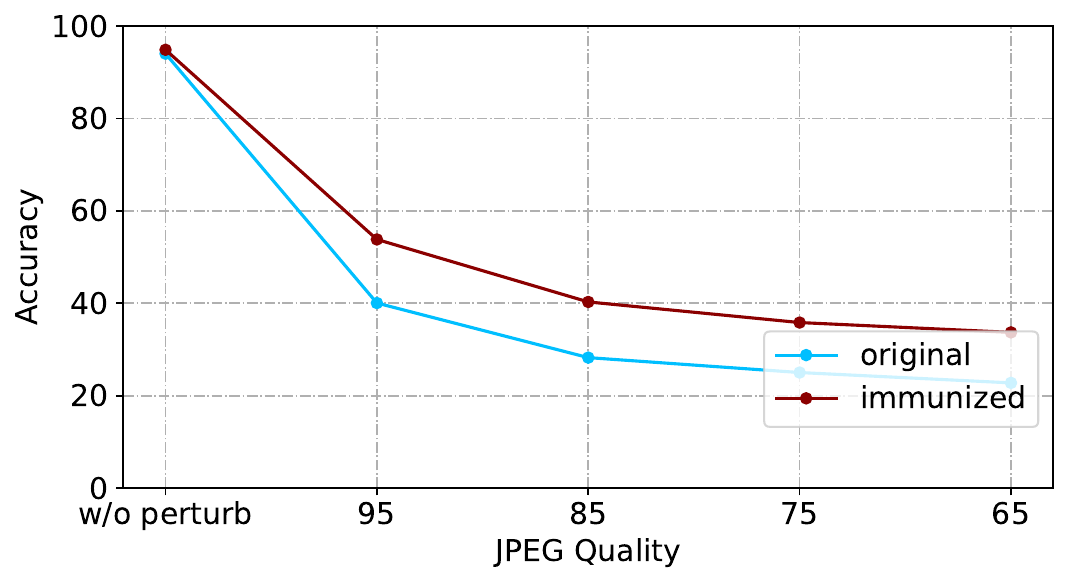}
\end{minipage}
\begin{minipage}{\linewidth}
\includegraphics[width=0.96\linewidth]{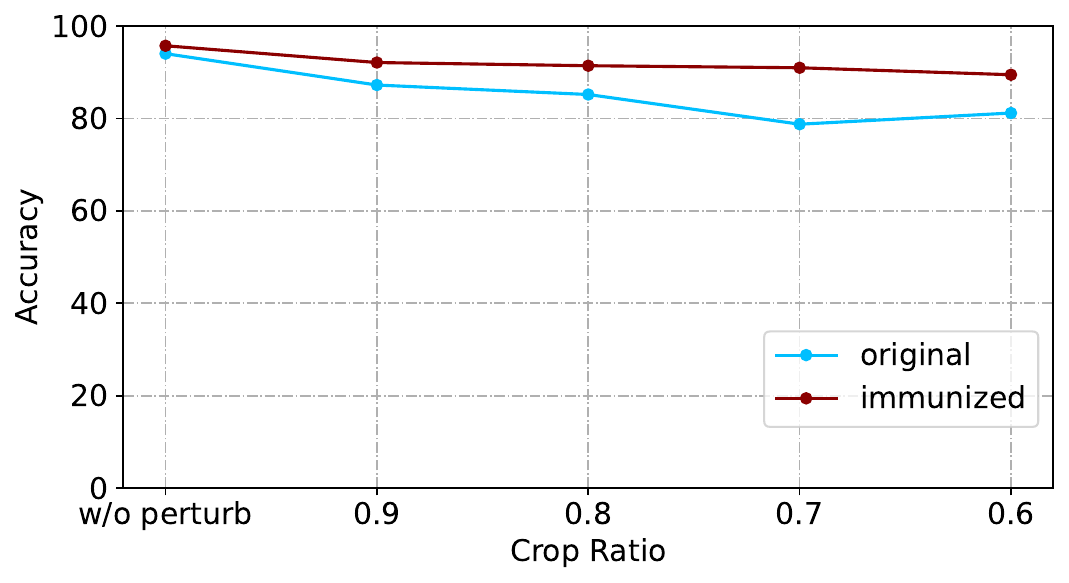}
\end{minipage}

\caption{Model identification accuracy of original and immunized fingerprint extractor under JPEG compression and crop resizing.} 
\label{fig:robust}
\end{figure}

\noindent \textbf{Image Perturbations.} Images generated by models may undergo various post-processings during transmission on social media. To evaluate the robustness of our fingerprint extractor (Only-Syn), we tested it against the two most common types of post-processing: JPEG compression and crop resizing. For these tests, we utilized 'ImageCompression' and 'RandomResizedCrop' from the Albumentations library, where 'RandomResizedCrop' randomly crops a patch from the image and resizes it back to the original dimensions. As depicted in Figure~\ref{fig:robust}, we charted the model identification accuracy alongside the JPEG compression rate and crop ratio. It was observed that our trained fingerprint extractor is relatively robust to the crop resizing post-processing, yet exhibits sensitivity to JPEG compression. To mitigate this sensitivity, we incorporate them as a form of data augmentation during training. The immunized fingerprint extractor, as depicted by the red line in the results, demonstrates strong robustness against crop resizing and improvement in resilience against JPEG compression. However, the performance does not fully recover to that of the unaltered data under JPEG compression. This is primarily because JPEG compression can significantly alter the spectrum pattern, thereby degrading the original pattern left by the model, as also noted in~\cite{corvi2023intriguing}. Addressing the robustness against JPEG compression will be a focus of the future work. 

\subsection{Additional Ablation Study}
\label{sec:appendix_ablation}

\noindent \textbf{Parameter $N$ for FFD Metric Calculation.} We conduct an ablation study to examine the impact of parameter $N$ on the FFD evaluation criterion. FFD values are calculated under various conditions: All, w/o K, w/o L, w/o S, w/o U, w/o A, w/o N, and w/o seed, across different numbers of images ($N$) ranging from 50 to 500. We then compute the Pearson rank correlation between the FFD values at $N = 500$ and those at each tested $N$. As shown in~\cref{tab:ablation_N}, the Pearson rank correlation with the baseline ($N = 500$) remains consistently close to 1, indicating that changes in $N$ do not affect the ranking of results. This demonstrates that the FFD metric is robust to variations in $N$ within the tested range.

\begin{table}[tbp]
\centering
\caption{Pearson rank correlation between the FFD values at $N$=500 and those across $N$ from 50 to 500.}
\label{tab:ablation_N}
\scalebox{0.98}{
\begin{tabular}{cccc}
\toprule[1pt]
Number of Images & Pearson Correlation with N=500\\  
\midrule
N=50 & 99.97 \%\\
N=100 & 99.98\%\\
N=200 &  99.99\% \\
N=300 &  100.00 \%\\
N=400 & 100.00 \%\\
N=500 & 100.00 \%\\
\bottomrule[1pt]
\end{tabular}}
\end{table}

\subsection{Additional Visualization Results.}

\noindent \textbf{Spectrum Visualization of Real and Synthetic Models.} In Figure~\ref{fig:syn_spectrum_compare}, we present additional spectrum visualizations of real models and their closest synthetic counterparts. The real models include GANs, VAEs, Flows, and Diffusion models. As shown, while the spectra of synthetic models do not perfectly match those of real models, they exhibit notable similarities and diverse frequency artifacts. These similarities provide empirical evidence that our synthetic training strategy effectively approximates the frequency characteristics of unseen models, thereby improving generalization.

\begin{figure*}[h]
\setlength{\abovecaptionskip}{0mm}
\centering

\includegraphics[width=0.7\linewidth]{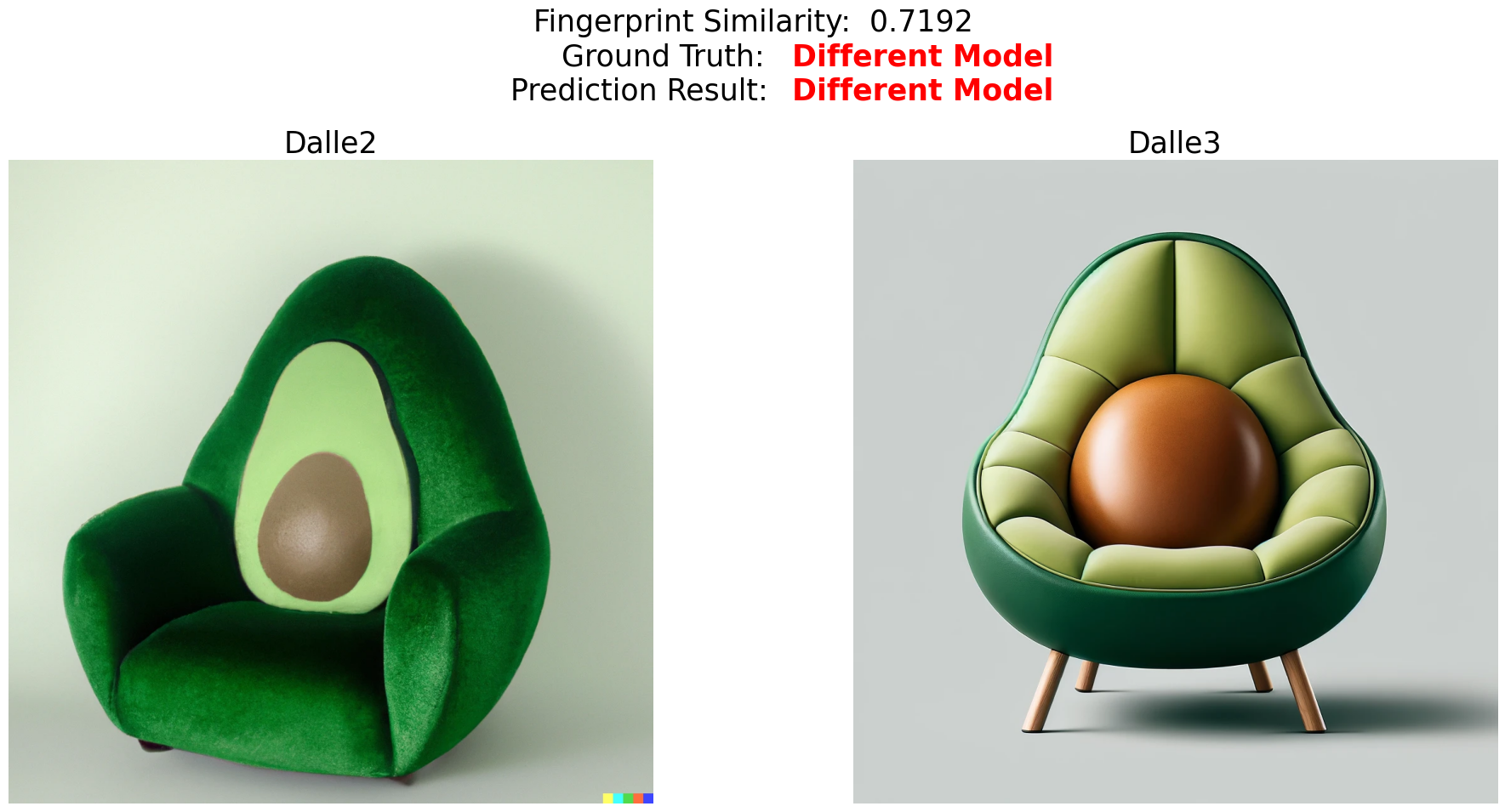} 
\begin{minipage}{0.4\linewidth}
    \centering \figtext{Image from the showcases on the DallE-2 webpage. \href{https://labs.openai.com/}{Link}\\
    Prompt: An armchair in the shape of an avocado.}
\end{minipage}
\begin{minipage}{0.4\linewidth}
    \centering \figtext{Generated by DallE-3 in ChatGPT. \\
    Prompt: An armchair in the shape of an avocado.}
\end{minipage}
\vspace{20pt}

\includegraphics[width=0.7\linewidth]{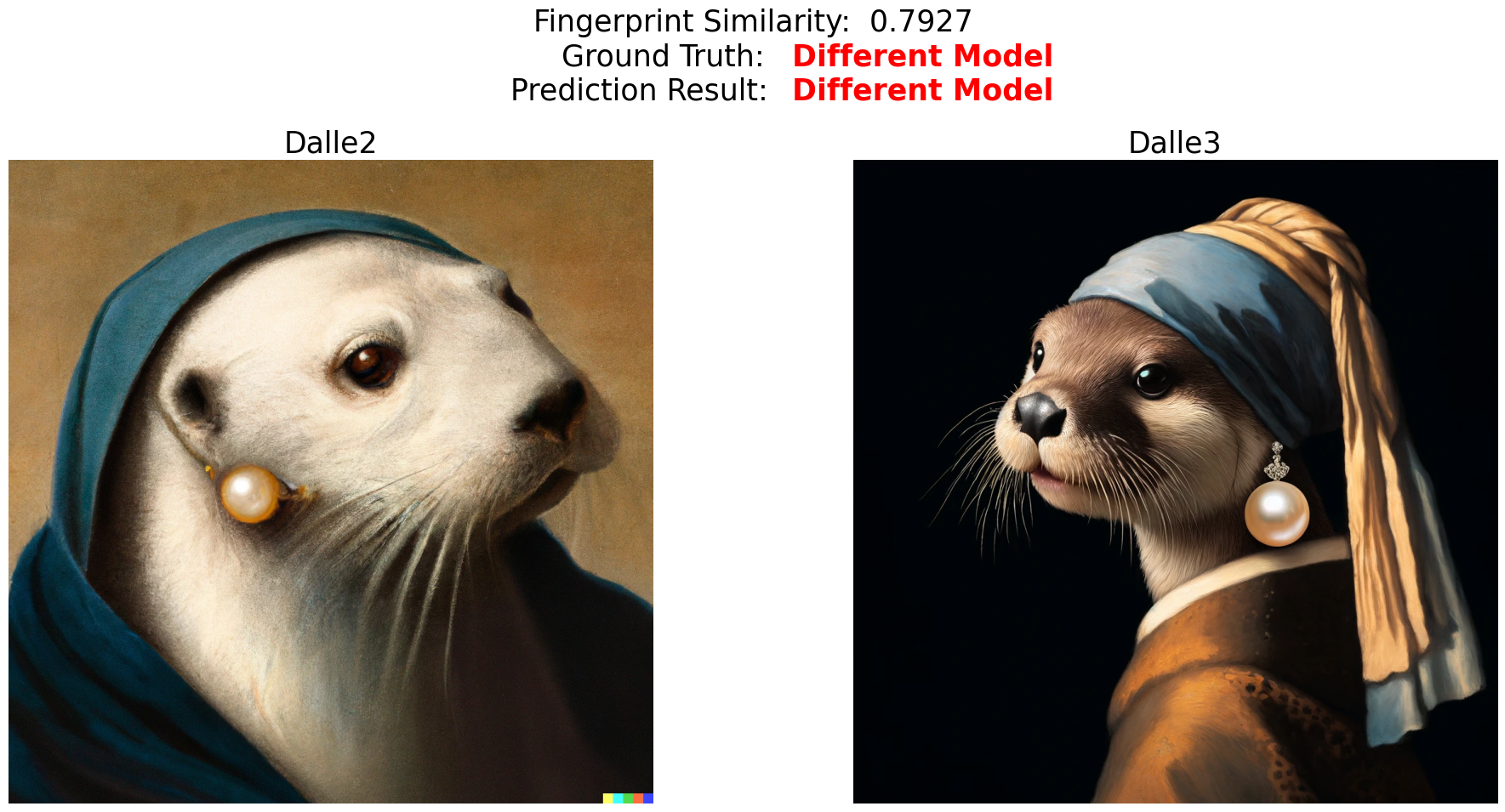} 
\begin{minipage}{0.4\linewidth}
    \centering \figtext{Image from the showcases on the DallE-2 webpage. \href{https://labs.openai.com/}{Link}\\
    Prompt: "A sea otter with a pearl earring" by Johannes Vermeer.}
\end{minipage}
\begin{minipage}{0.4\linewidth}
    \centering \figtext{Generated by DallE-3 in ChatGPT. \\ Prompt: "A sea otter with a pearl earring" by Johannes Vermeer.}
\end{minipage}
\vspace{20pt}

\includegraphics[width=0.7\linewidth]{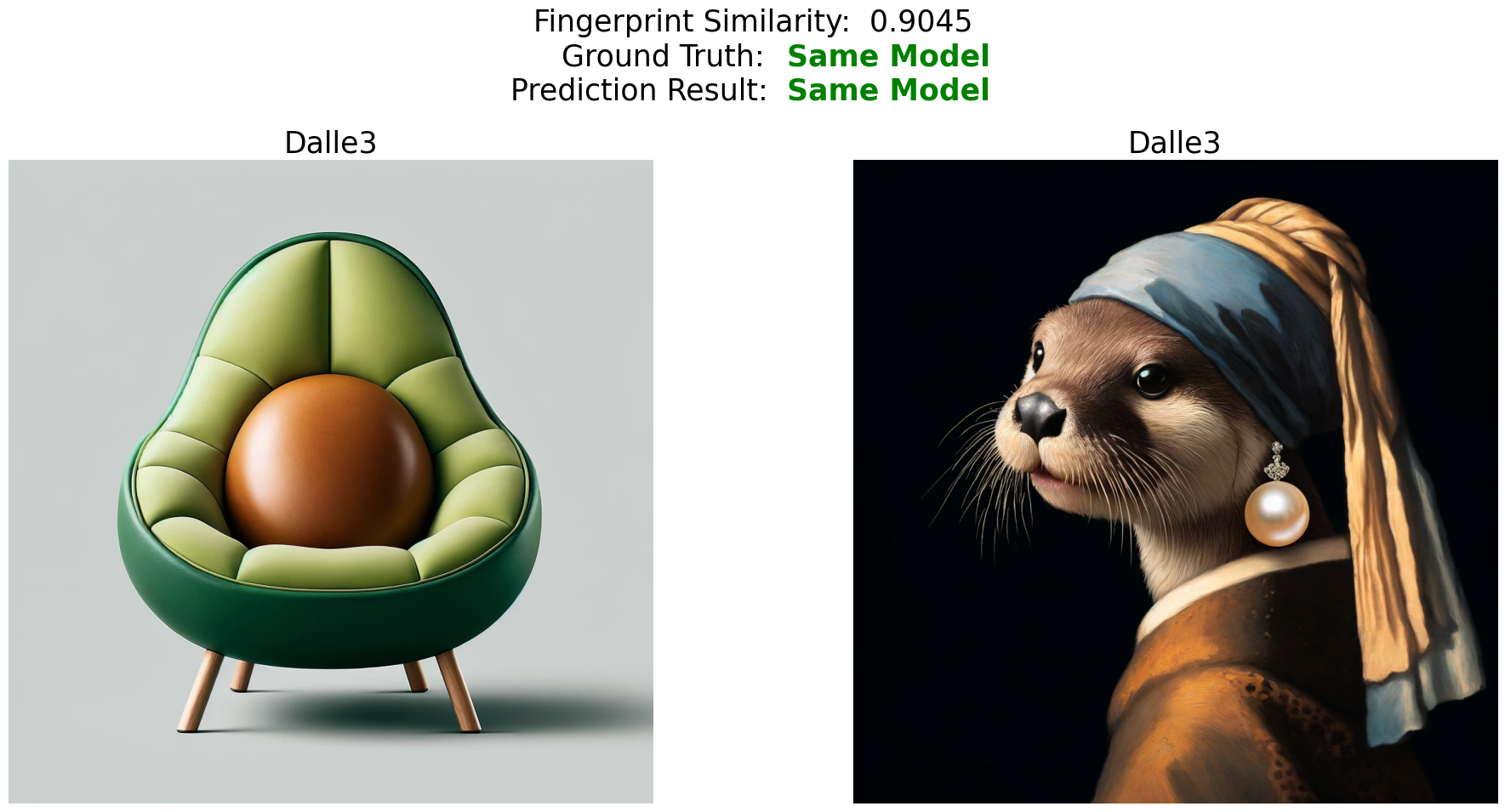} 
\begin{minipage}{0.4\linewidth}
    \centering \figtext{Generated by DallE-3 in ChatGPT. \\ Prompt: An armchair in the shape of an avocado}
\end{minipage}
\begin{minipage}{0.4\linewidth}
    \centering \figtext{Generated by DallE-3 in ChatGPT. \\ Prompt: "A sea otter with a pearl earring" by Johannes Vermeer}
\end{minipage}
\vspace{20pt}

\caption{Model verification results on images from DallE-2 and DallE-3.}
\label{fig:cases_dalle}
\end{figure*}

\begin{figure*}[h!]
\centering
\includegraphics[width=0.7\linewidth]{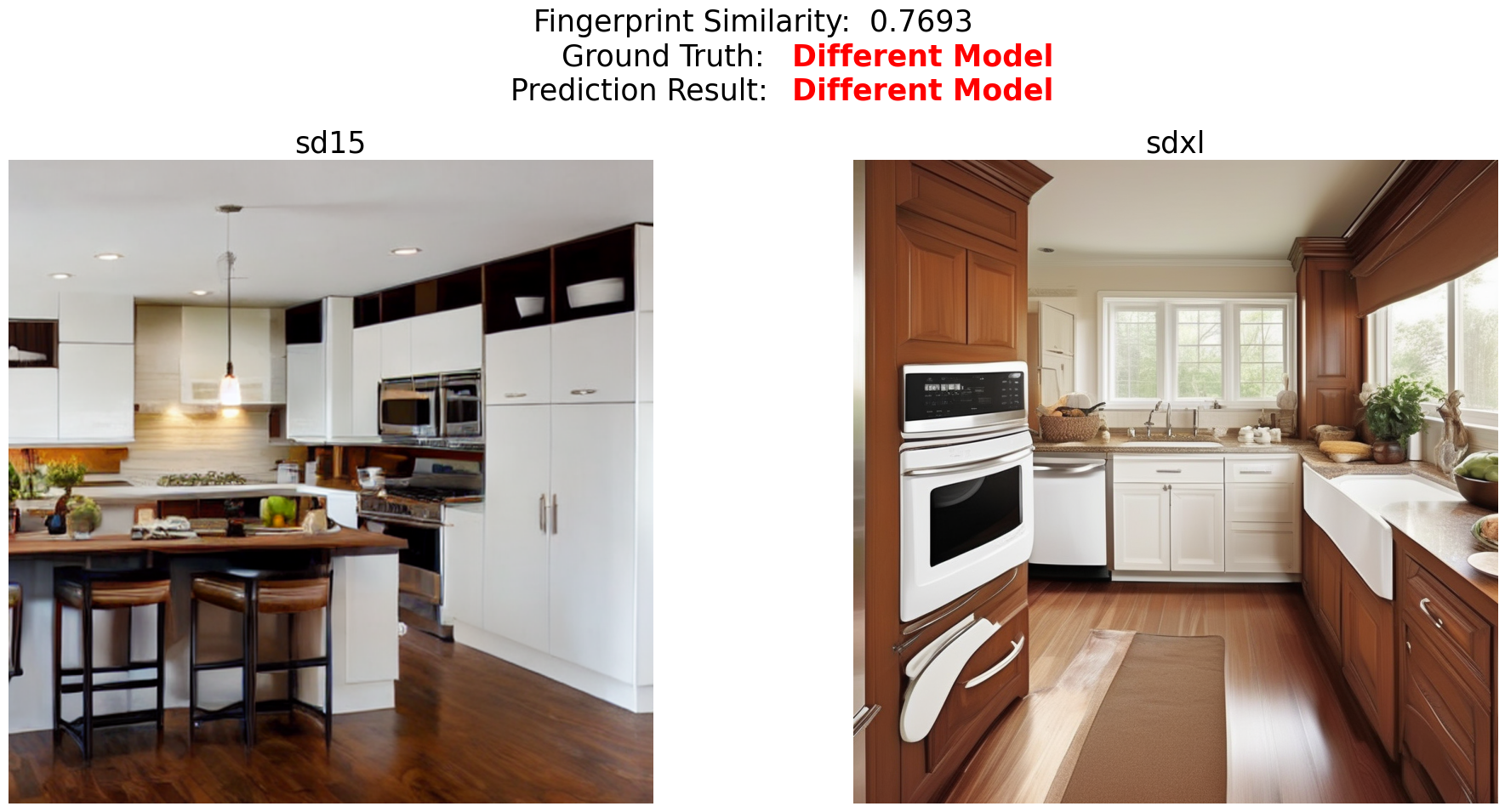} 
\begin{minipage}{0.4\linewidth}
    \centering \figtext{Prompt: Stark white appliances stand out against brown wooden cabinets.}
\end{minipage}
\begin{minipage}{0.4\linewidth}
    \centering \figtext{Prompt: Stark white appliances stand out against brown wooden cabinets.}
\end{minipage}
\vspace{20pt}

\includegraphics[width=0.7\linewidth]{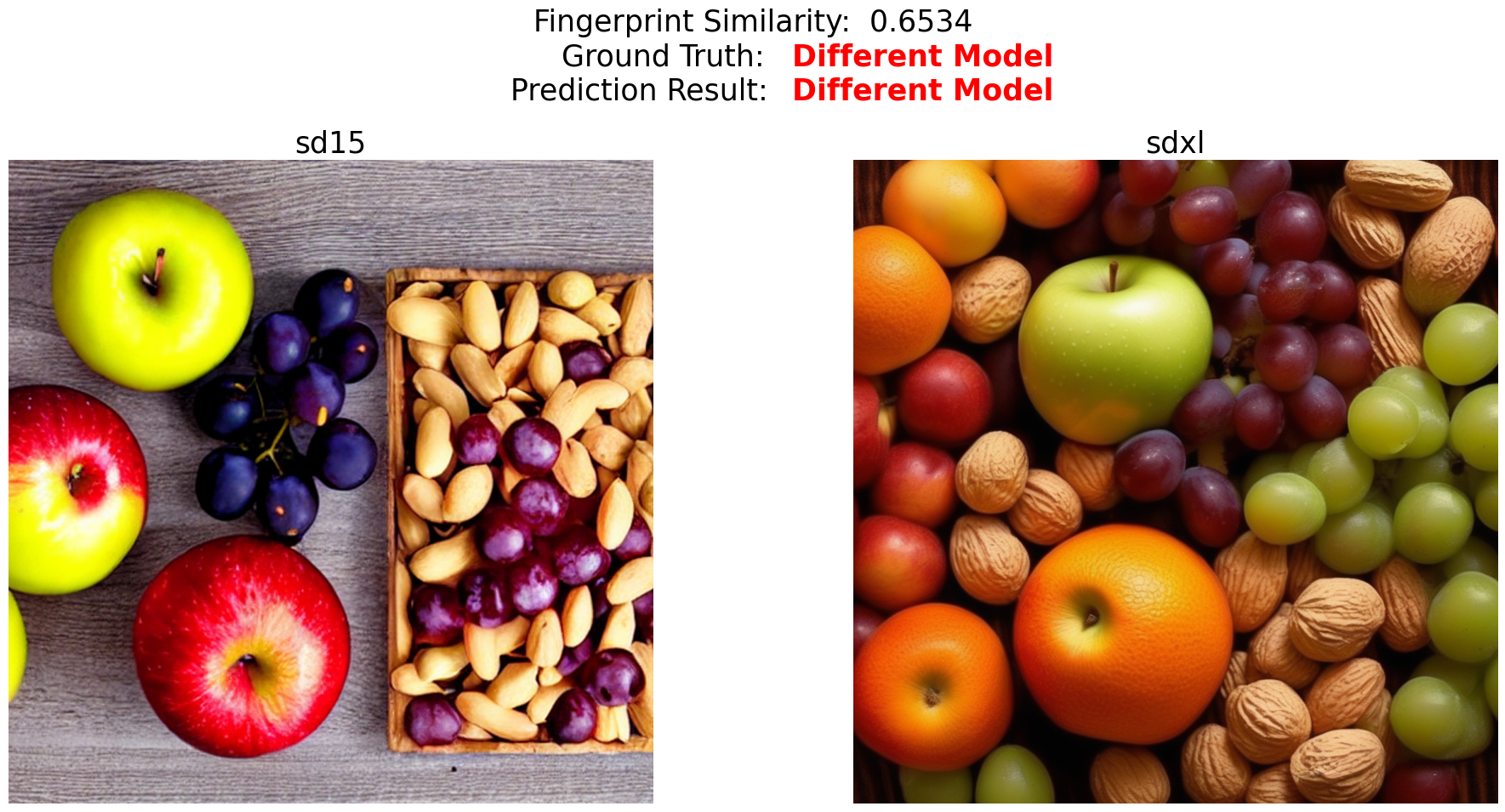} 
\begin{minipage}{0.4\linewidth}
    \centering \figtext{Prompt: Two apples, an orange, some grapes and peanuts.}
\end{minipage}
\begin{minipage}{0.4\linewidth}
    \centering \figtext{Prompt: Two apples, an orange, some grapes and peanuts.}
\end{minipage}
\vspace{20pt}

\includegraphics[width=0.7\linewidth]{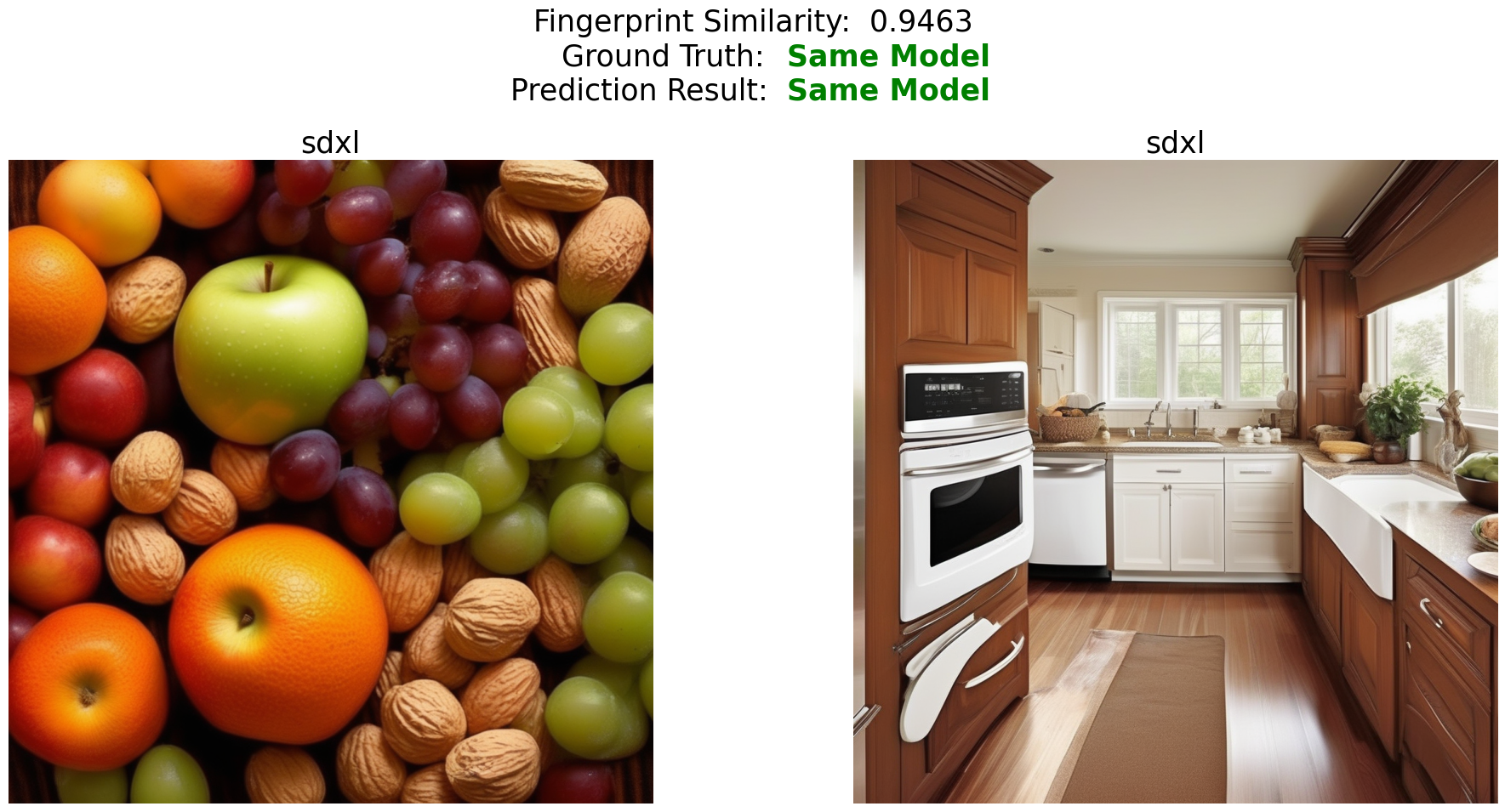} 
\begin{minipage}{0.4\linewidth}
    \centering \figtext{Prompt: Two apples, an orange, some grapes and peanuts.}
\end{minipage}
\begin{minipage}{0.4\linewidth}
    \centering \figtext{Prompt: Stark white appliances stand out against brown wooden cabinets.}
\end{minipage}
\vspace{20pt}

\caption{Model verification results on images from Stable Diffusion models.}
\label{fig:cases_sd}

\end{figure*}

\begin{figure*}[h]
\setlength{\abovecaptionskip}{0mm}
\centering
\includegraphics[width=0.7\linewidth]{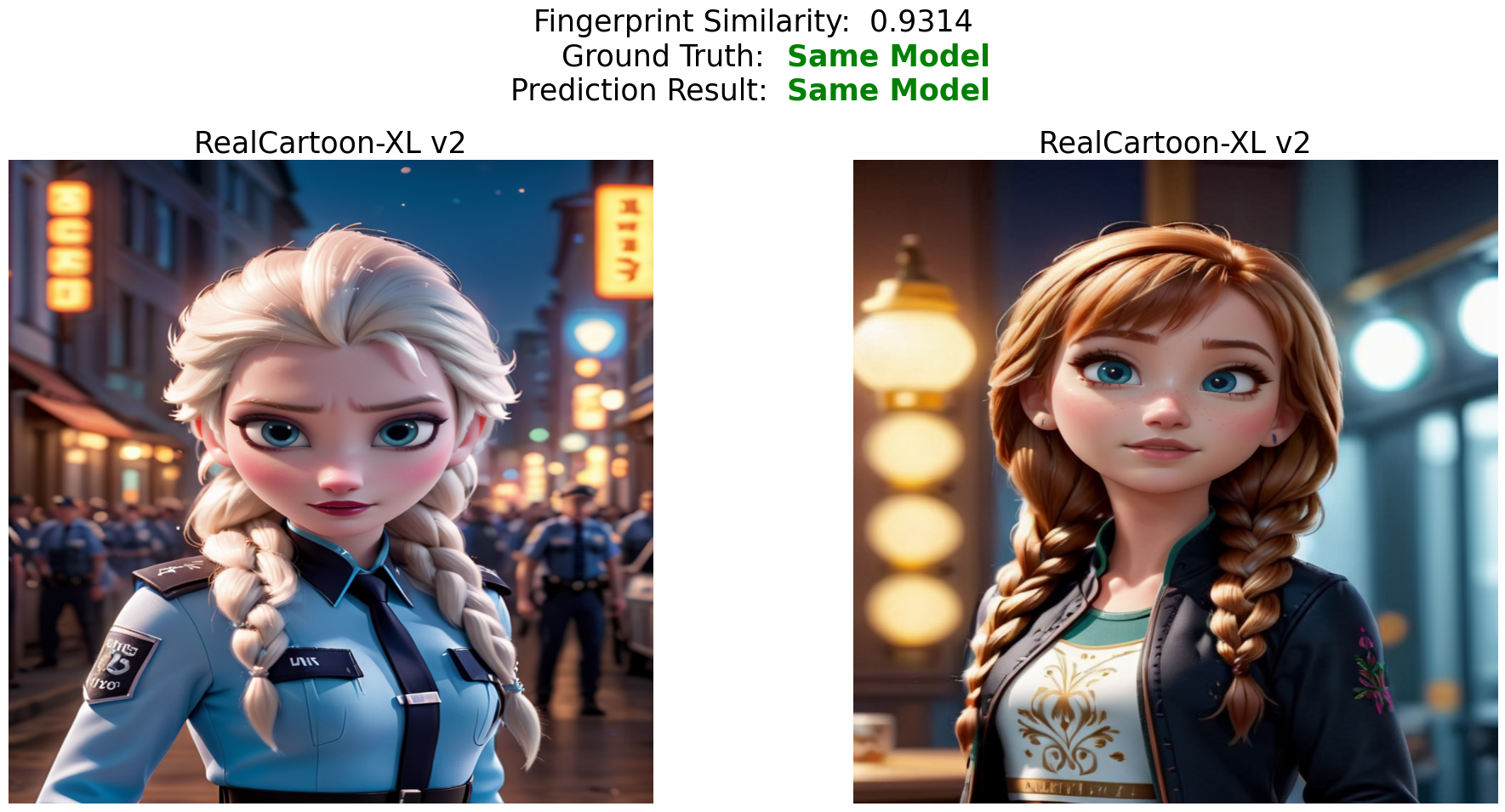} 
\begin{minipage}{0.4\linewidth}
    \centering \figtext{Image from https://civitai.com.
    \\\href{https://civitai.com/models/212532/all-disney-princess-xl-lora-model-from-ralph-breaks-the-internet}{Link}}
\end{minipage}
\begin{minipage}{0.4\linewidth}
    \centering \figtext{Image from https://civitai.com. \\\href{https://civitai.com/models/212532/all-disney-princess-xl-lora-model-from-ralph-breaks-the-internet}{Link}}
\end{minipage}
\vspace{20pt}

\includegraphics[width=0.7\linewidth]{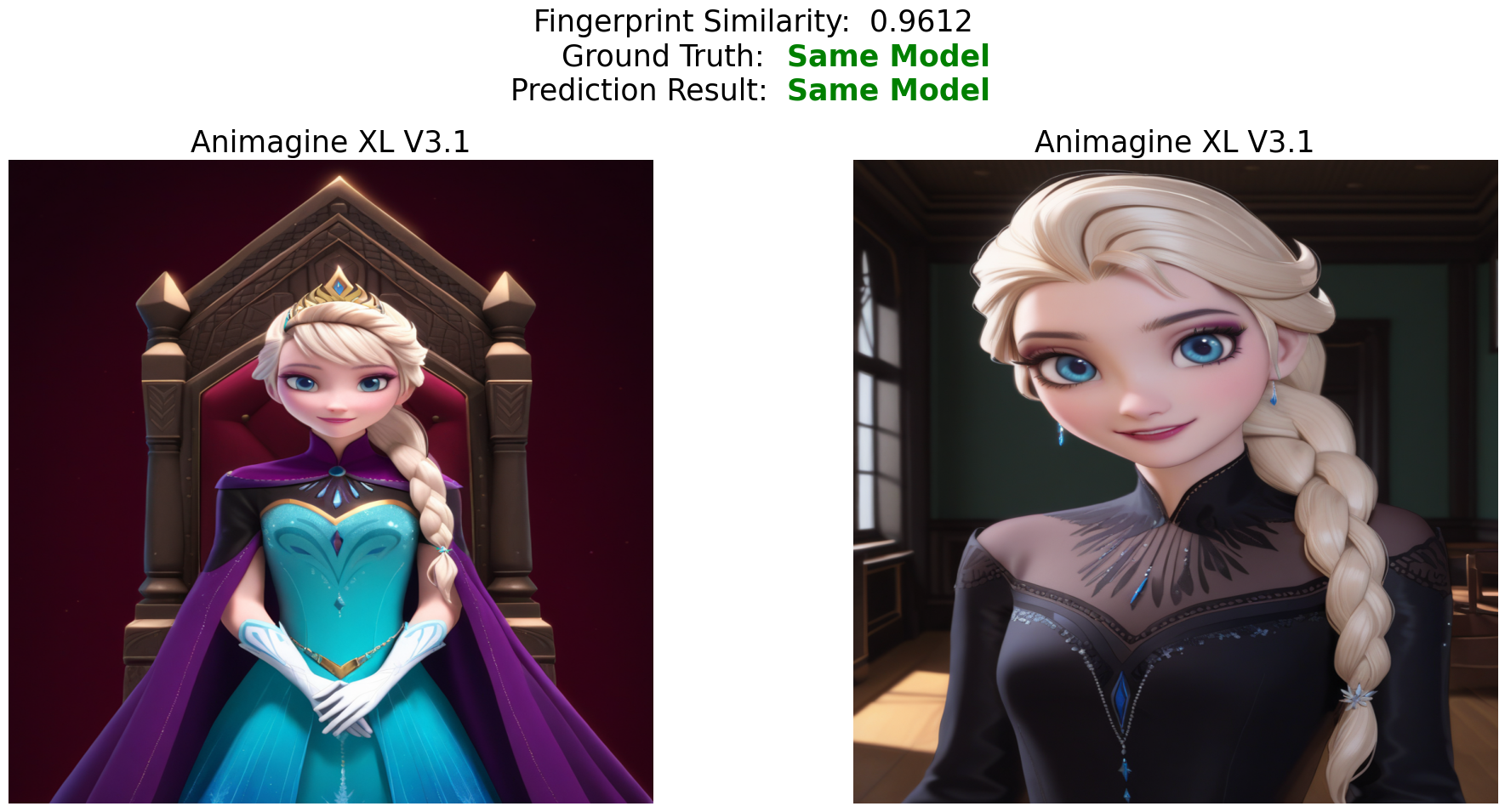} 
\begin{minipage}{0.4\linewidth}
    \centering \figtext{Image from https://civitai.com.
    \\\href{https://civitai.com/models/332276/elsa-frozen-sd15sdxlponysdxl-commission}{Link}}
\end{minipage}
\begin{minipage}{0.4\linewidth}
    \centering \figtext{Image from https://civitai.com.
    \\\href{https://civitai.com/models/332276/elsa-frozen-sd15sdxlponysdxl-commission}{Link}}
\end{minipage}
\vspace{20pt}

\includegraphics[width=0.7\linewidth]{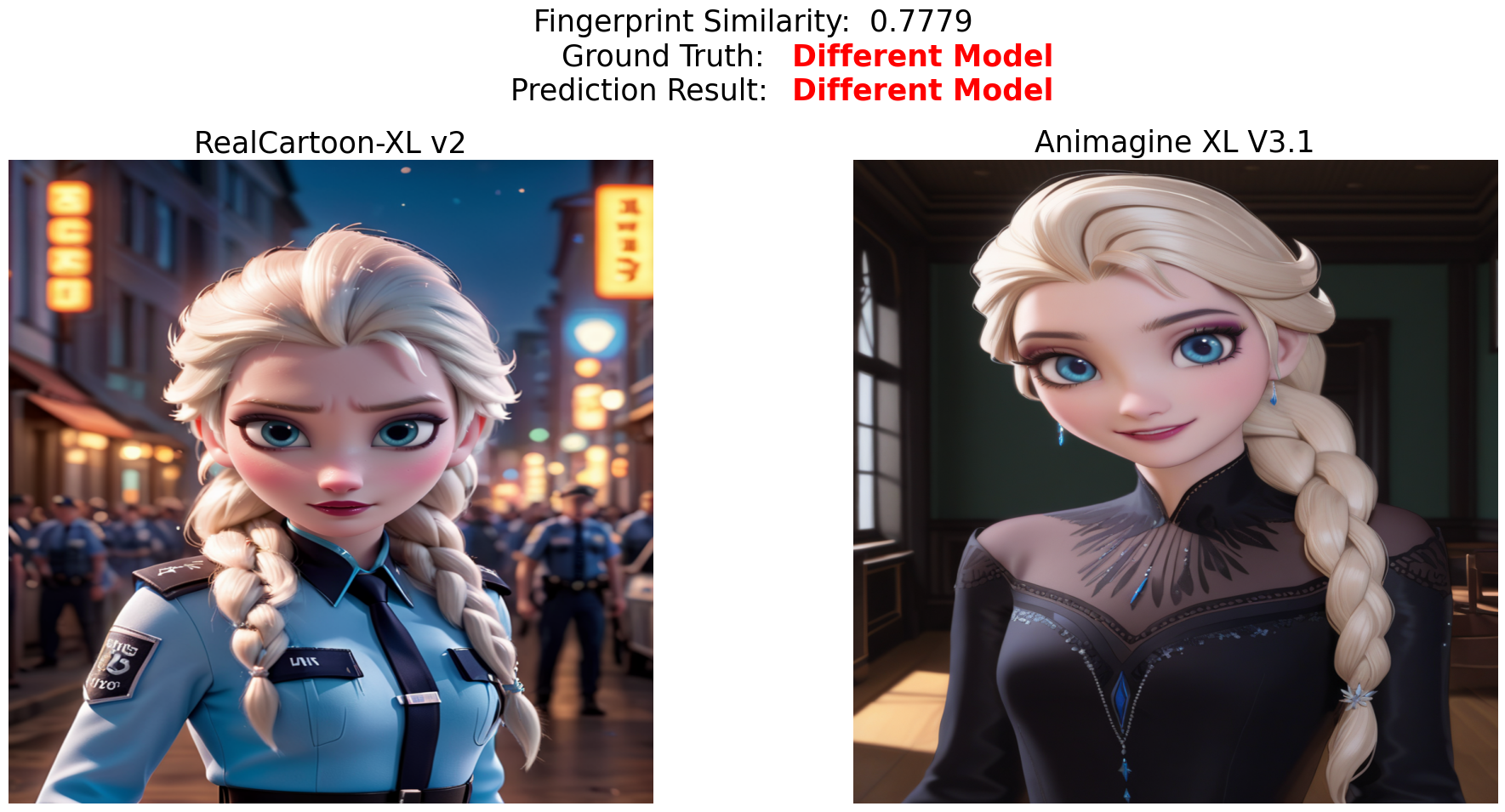} 
\begin{minipage}{0.4\linewidth}
    \centering \figtext{Image from https://civitai.com.
    \\\href{https://civitai.com/models/212532/all-disney-princess-xl-lora-model-from-ralph-breaks-the-internet}{Link}}
\end{minipage}
\begin{minipage}{0.4\linewidth}
    \centering \figtext{Image from https://civitai.com.
    \\\href{https://civitai.com/models/332276/elsa-frozen-sd15sdxlponysdxl-commission}{Link}}
\end{minipage}
\vspace{20pt}

\caption{Model verification results on images from https://civitai.com.}
\label{fig:cases_civitai}
\end{figure*}

\noindent \textbf{Model Verification Cases.} We provide some model verification cases produced by our model fingerprint extractor in Figure~\ref{fig:cases_dalle}, Figure~\ref{fig:cases_sd}, and Figure~\ref{fig:cases_civitai}.  Model verification aims at verifying whether two generated images are from the same model by comparing extracted fingerprints. The threshold to make the prediction is empirically set as 0.81, which is the best threshold of model verification on the Text2Image models in Table~\ref{tab:dataset}. \textit{It is important to note that, to evaluate the zero-shot attribution ability in the open world, the fingerprint extractor producing these results is solely trained on synthetic models, without having seen any images from the real generative models tested below.}

\begin{figure*}
\centering
\begin{minipage}{0.2\linewidth}
    \centering
    \figtext{\textbf{Real Models}}
\end{minipage}
\begin{minipage}{0.2\linewidth}
    \centering
    \figtext{\textbf{Synthetic Models}}
\end{minipage}
\begin{minipage}{0.2\linewidth}
    \centering
    \figtext{\textbf{Real Models}}
\end{minipage}
\begin{minipage}{0.2\linewidth}
    \centering
    \figtext{\textbf{Synthetic Models}}
\end{minipage}

\begin{minipage}{\subfigurewidth\linewidth}
    \includegraphics[width=\linewidth]{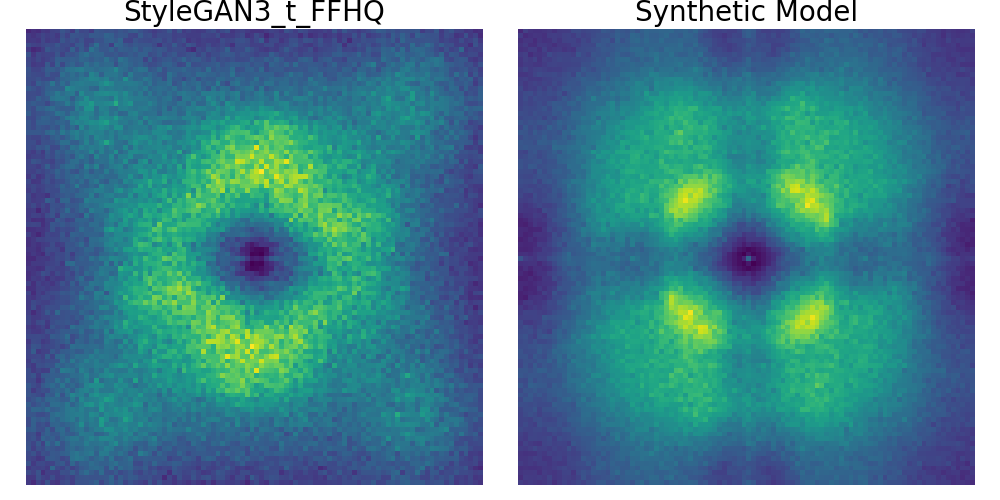}
\end{minipage}
\begin{minipage}{\subfigurewidth\linewidth}
    \includegraphics[width=\linewidth]{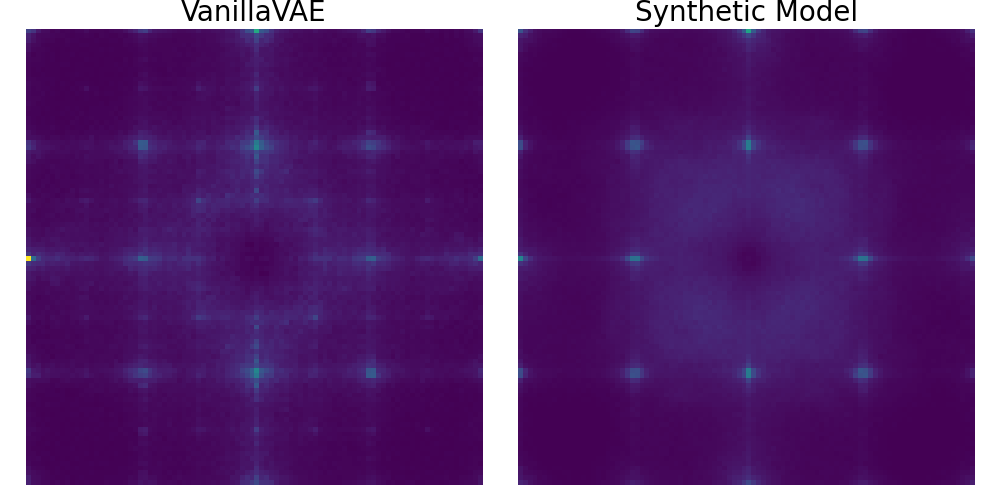}
\end{minipage}
\begin{minipage}{\subfigurewidth\linewidth}
    \includegraphics[width=\linewidth]{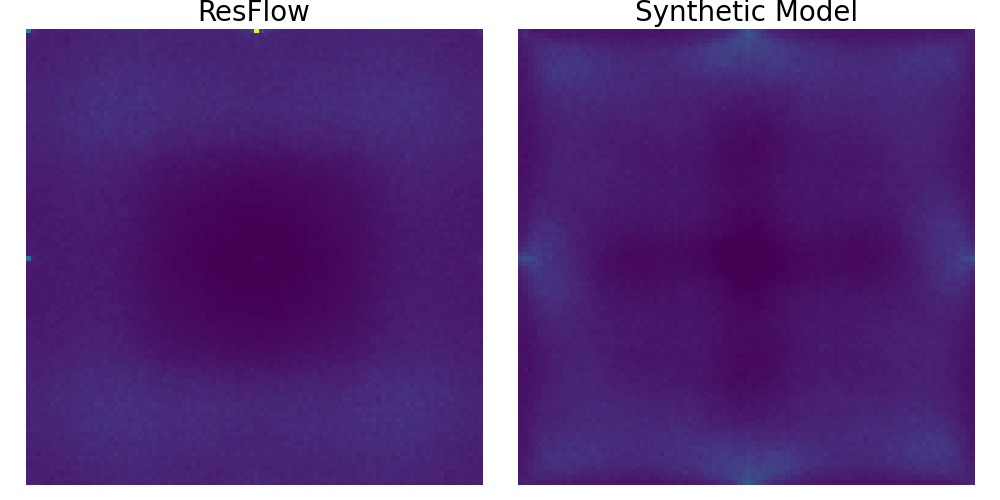}
\end{minipage}
\begin{minipage}{\subfigurewidth\linewidth}
    \includegraphics[width=\linewidth]{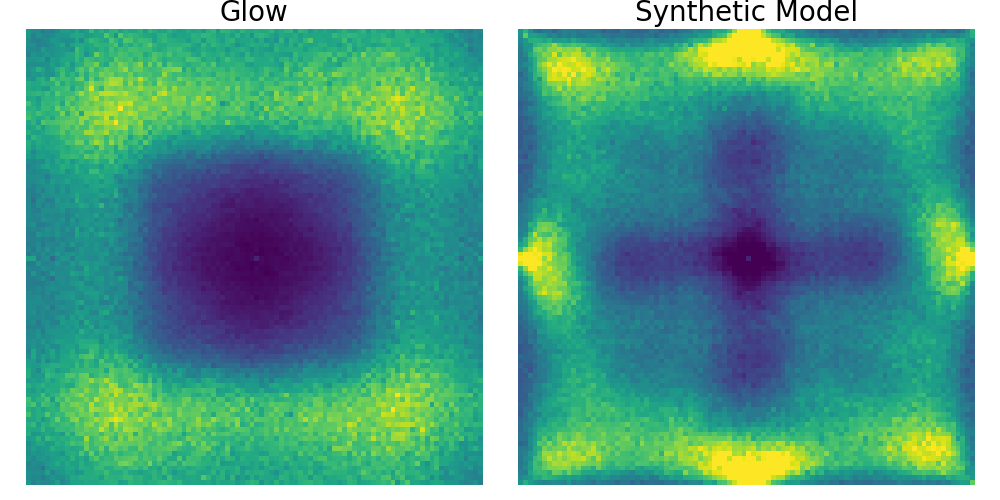}
\end{minipage}

\begin{minipage}{\subfigurewidth\linewidth}
    \includegraphics[width=\linewidth]{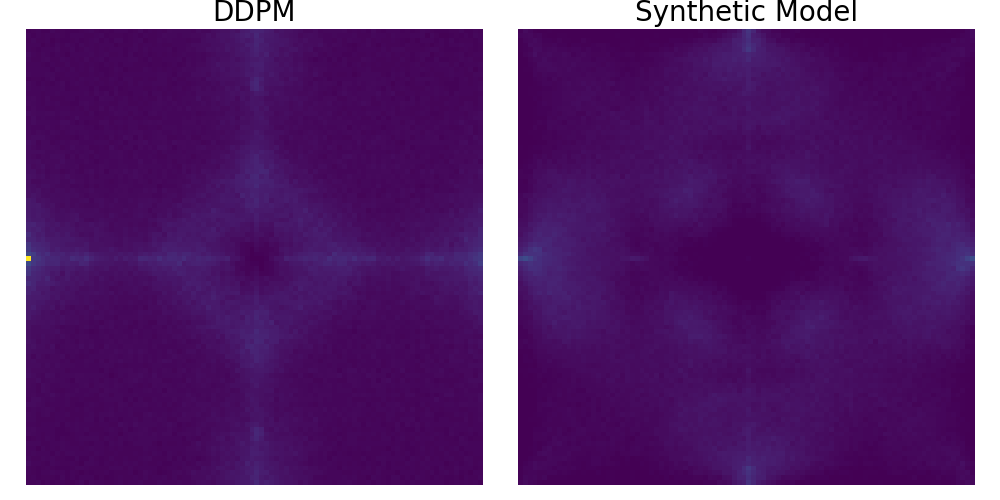}
\end{minipage}
\begin{minipage}{\subfigurewidth\linewidth}
    \includegraphics[width=\linewidth]{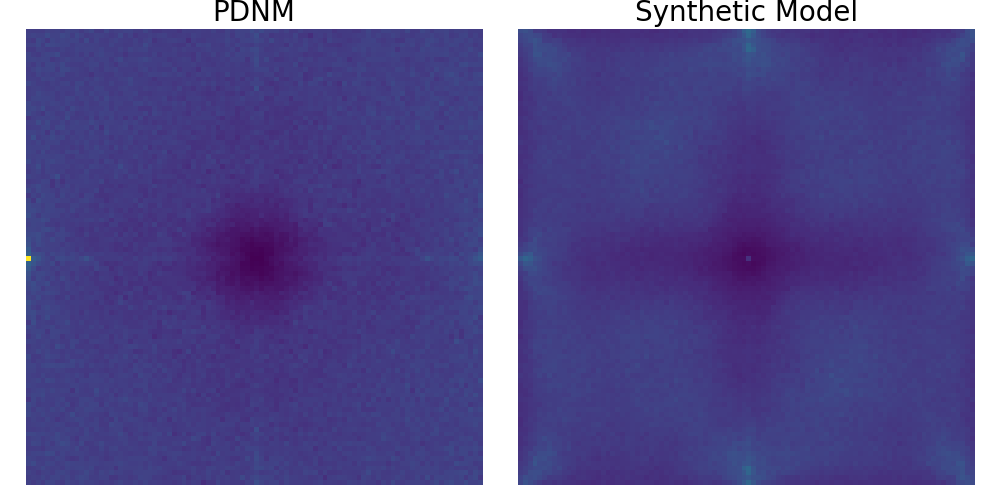}
\end{minipage}

\begin{minipage}{\subfigurewidth\linewidth}
    \includegraphics[width=\linewidth]{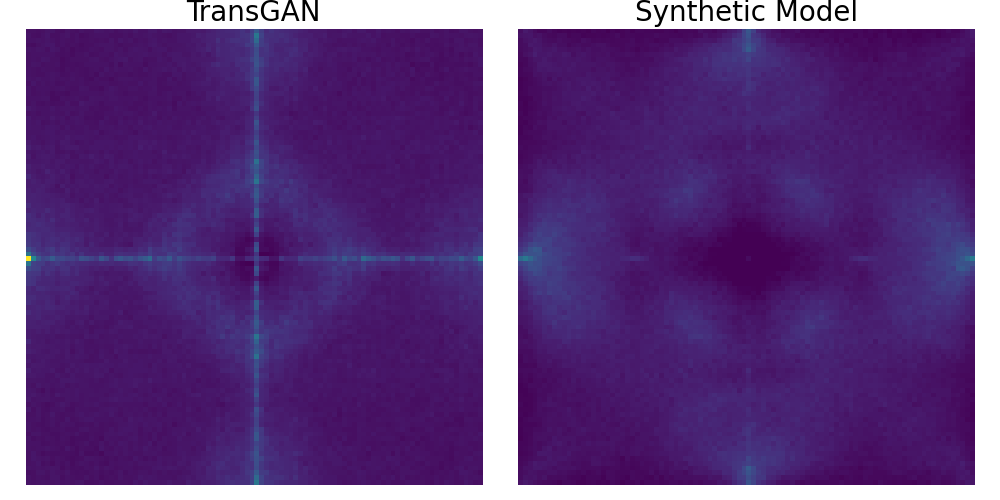}
\end{minipage}
\begin{minipage}{\subfigurewidth\linewidth}
    \includegraphics[width=\linewidth]{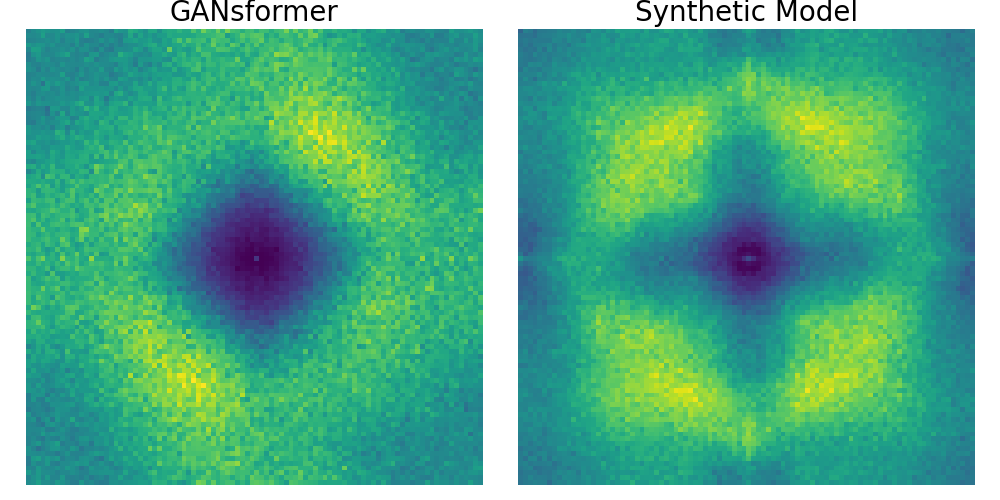}
\end{minipage}

\caption{Spectrum of real-world models and its closest synthetic model within the 5760 models. The synthetic model could mimic diverse spectrum and resemble with spectrum real world models.}
\end{figure*}

\begin{figure*}
\ContinuedFloat
\centering
\begin{minipage}{0.2\linewidth}
    \centering
    \figtext{\textbf{Real Models}}
\end{minipage}
\begin{minipage}{0.2\linewidth}
    \centering
    \figtext{\textbf{Synthetic Models}}
\end{minipage}
\begin{minipage}{0.2\linewidth}
    \centering
    \figtext{\textbf{Real Models}}
\end{minipage}
\begin{minipage}{0.2\linewidth}
    \centering
    \figtext{\textbf{Synthetic Models}}
\end{minipage}

\begin{minipage}{\subfigurewidth\linewidth}
    \includegraphics[width=\linewidth]{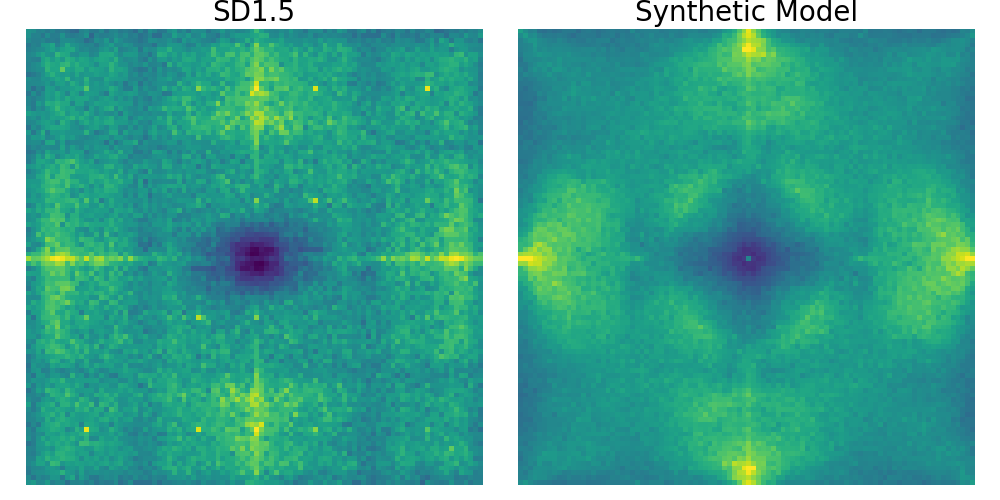}
\end{minipage}
\begin{minipage}{\subfigurewidth\linewidth}
    \includegraphics[width=\linewidth]{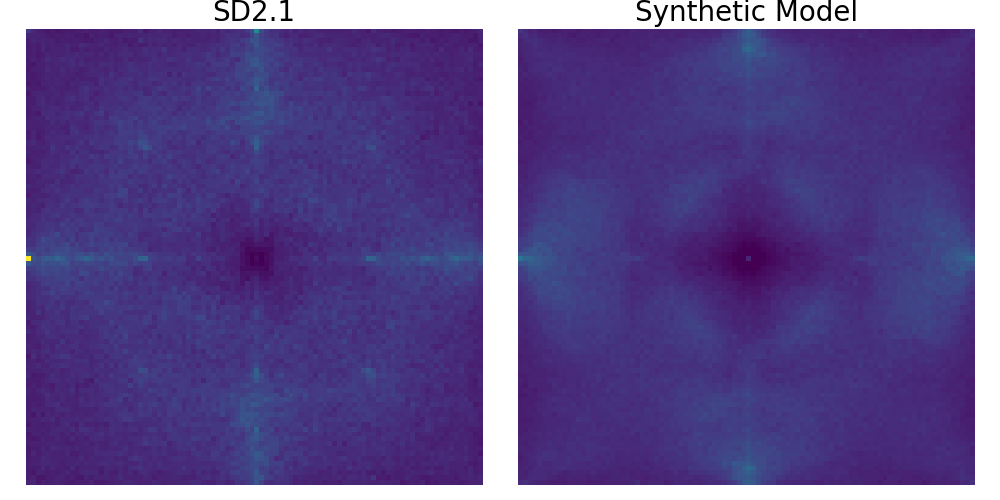}
\end{minipage}
\begin{minipage}{\subfigurewidth\linewidth}
    \includegraphics[width=\linewidth]{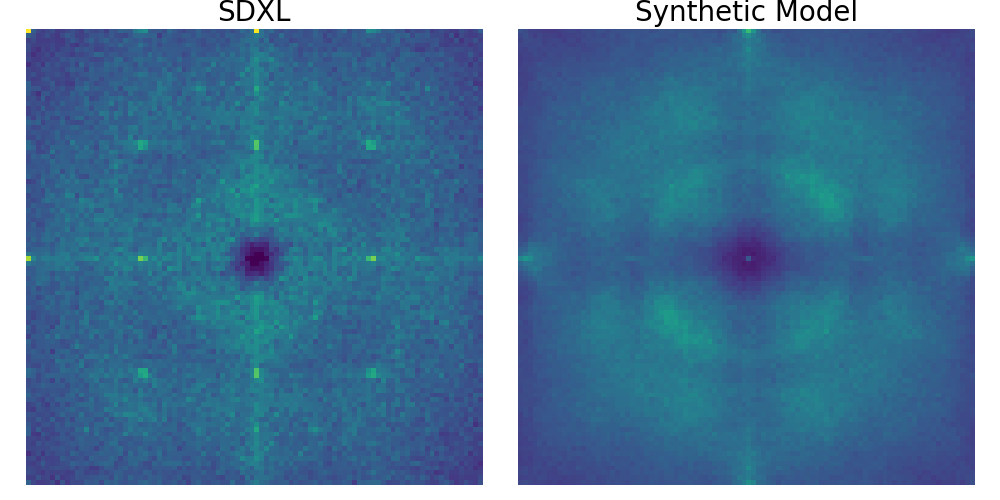}
\end{minipage}
\begin{minipage}{\subfigurewidth\linewidth}
    \includegraphics[width=\linewidth]{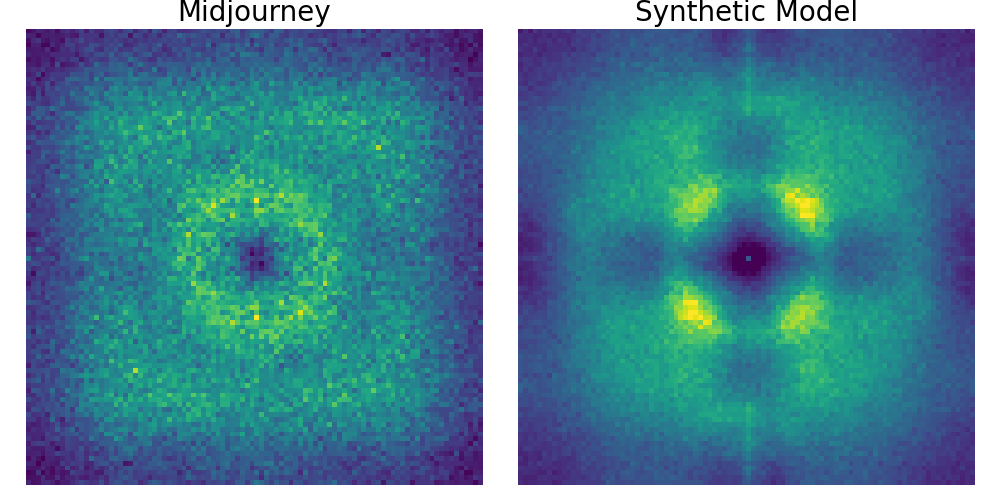}
\end{minipage}

\begin{minipage}{\subfigurewidth\linewidth}
    \includegraphics[width=\linewidth]{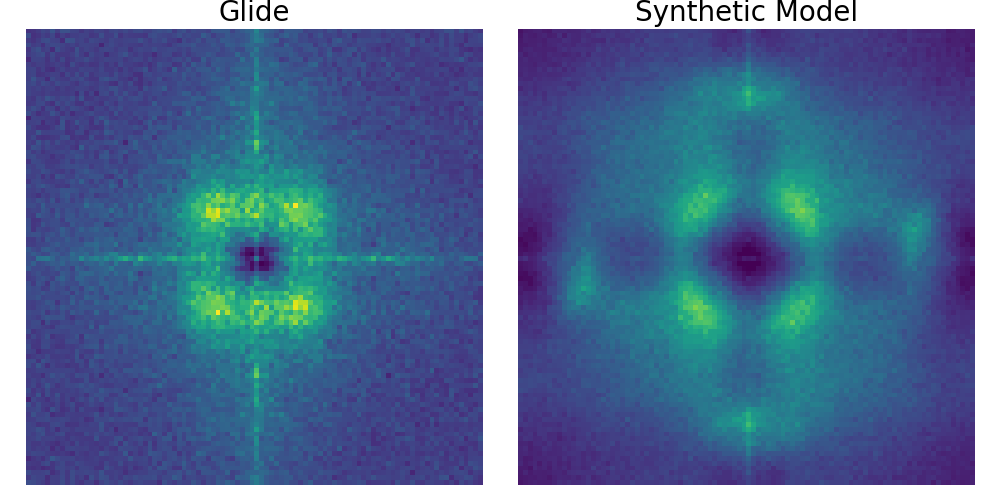}
\end{minipage}
\begin{minipage}{\subfigurewidth\linewidth}
    \includegraphics[width=\linewidth]{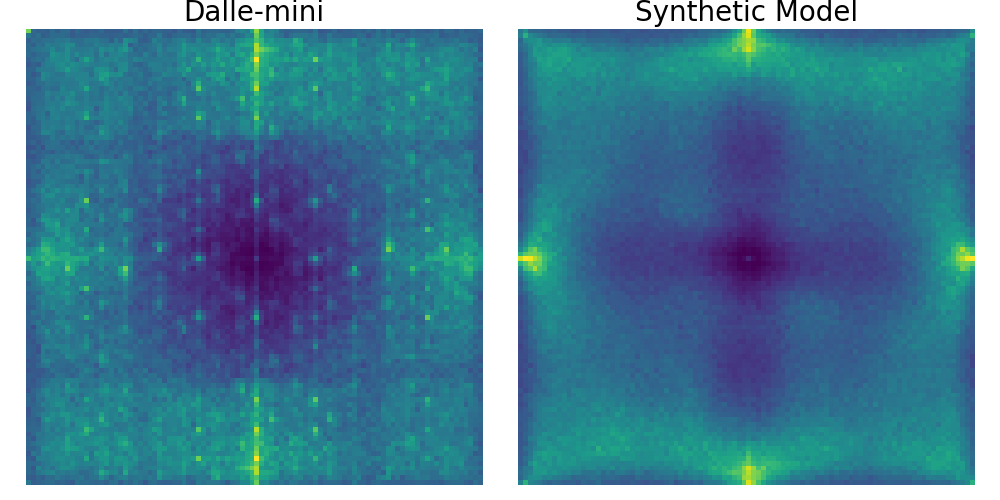}
\end{minipage}
\begin{minipage}{\subfigurewidth\linewidth}
    \includegraphics[width=\linewidth]{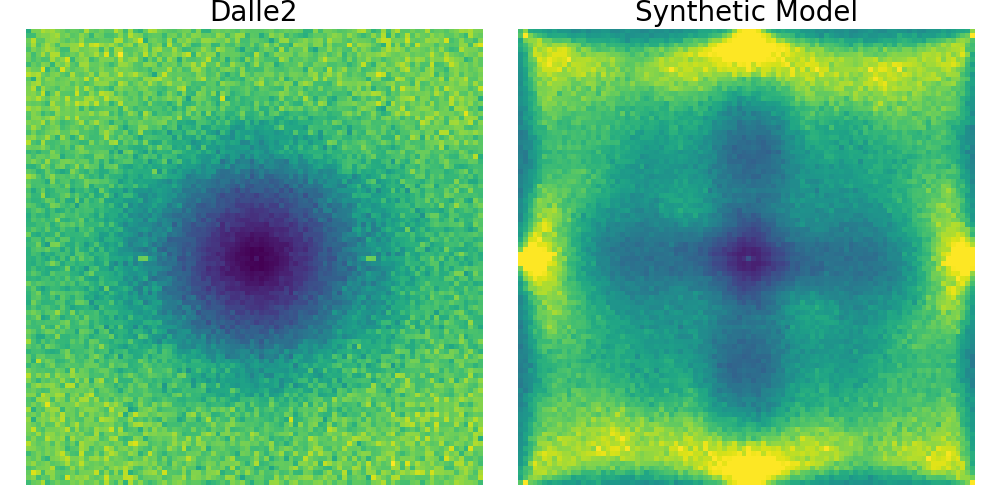}
\end{minipage}
\begin{minipage}{\subfigurewidth\linewidth}
    \includegraphics[width=\linewidth]{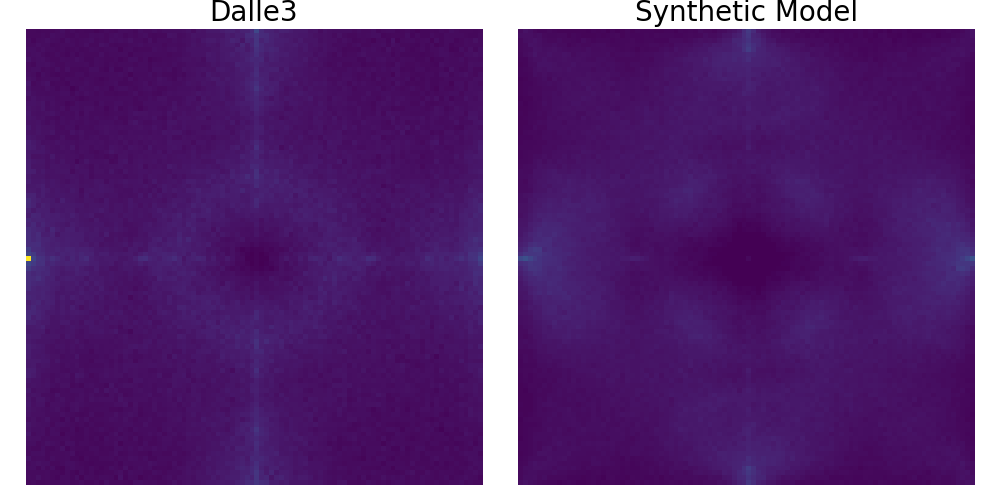}
\end{minipage}

\caption{Spectrum of real-world models and its closest synthetic model within the 5760 models. The synthetic model could mimic diverse spectrum and resemble with spectrum real world models.\emph{(cont.)}}
\label{fig:syn_spectrum_compare}
\end{figure*}

\subsection{Experimental Setups}
When training on synthetic models, the real images used to generated fingerprinted images are firstly resized to 128$\times$128. The batch size is set to 200. The initial learning rate is $1e^{-4}$ with a warm up cosine scheduler. The margin for the triplet loss is 0.3. The total training epoch is 40. 

\subsection{Limitations}

As depicted in Figure~\ref{fig:cvitai_lora} in the main text, the extracted fingerprints from images of LoRA-equiped models are similar to their Base models, which allows us to attribute the LoRA variants to their base models. However, this also indicates a limitation of our method in distinguishing different LoRA variants from the same base model. We leave this more fine-grained model attribution challenge for future work.

\end{document}